\documentclass[12pt,a4paper,reqno]{article}
\usepackage{geometry}
\usepackage[font=footnotesize,labelfont=bf]{caption}

\usepackage[htt]{hyphenat}
\usepackage{amsmath}
\usepackage{amsfonts}
\usepackage{amssymb}
\usepackage{amsthm}

\usepackage{standalone}

\usepackage{mathtools}

\usepackage{xcolor}
\definecolor{morange}{RGB}{255,127,14}
\definecolor{mblue}{RGB}{31,119,180}
\definecolor{mred}{RGB}{214,39,40}
\definecolor{mpurple}{RGB}{148,103,189}
\definecolor{mgreen}{RGB}{44,160,44}
\usepackage{pifont}
\newcommand{\cmark}{\textcolor{mgreen}{\ding{51}}}
\newcommand{\xmark}{\textcolor{mred}{\ding{55}}}
\usepackage{graphicx}
\usepackage[unicode,bookmarks,colorlinks=true,breaklinks=true]{hyperref}
\hypersetup{
    pdfpagelabels,
    bookmarks,
    hyperindex,
    unicode = true,
    linkcolor = mred,
    urlcolor = mpurple,
    colorlinks = true,
    citecolor = mblue,
    menucolor = mred,
}
\usepackage{booktabs}
\newcommand{\botrule}{\bottomrule}
\usepackage[utf8]{inputenc}
\usepackage[numbers,square]{natbib}
\usepackage{stmaryrd}
\usepackage{tikz}

\usepackage{algorithm}
\usepackage{algpseudocode}
\makeatletter
\renewcommand{\ALG@beginalgorithmic}{\small}
\makeatother

\usepackage{hyperref}

\usepackage{appendix}

\newtheorem{theorem}{Theorem}
\newtheorem{proposition}{Proposition}
\newtheorem{definition}{Definition}
\newtheorem{operation}{Operation}
\newtheorem*{remark}{Remark}

\usepackage{tikz}
\usetikzlibrary{positioning}
\usetikzlibrary{matrix}
\usetikzlibrary{fit}
\usetikzlibrary{arrows.meta}
\usetikzlibrary{shapes.geometric}

\usepackage{nicematrix}

\usepackage{comment}

\usepackage{subcaption}

\usepackage{cleveref}
\crefname{operation}{operation}{operations}

\usepackage{orcidlink}

\begin{document}

\title{Generalizing Adam To Manifolds For Efficiently Training Transformers}

\author{
\large{Benedikt Brantner}\\
\small{(benedikt.brantner@ipp.mpg.de)}\orcidlink{0009-0001-7067-4612}
\vspace{.5em}\\
\normalsize{Max-Planck-Institut f\"ur Plasmaphysik}\\
\normalsize{Boltzmannstra\ss{}e 2, 85748 Garching, Deutschland}%
\vspace{.1em}\\
\small{and}
\vspace{.1em}\\
\normalsize{Technische Universit\"at M\"unchen, Zentrum Mathematik}\\
\normalsize{Boltzmannstra\ss{}e 3, 85748 Garching, Deutschland}%
\vspace{1em}\\
}

\date{\today}

\maketitle

\begin{abstract}
One of the primary reasons behind the success of neural networks has been the emergence of an array of new, highly-successful optimizers, perhaps most importantly the Adam optimizer. It is widely used for training neural networks, yet notoriously hard to interpret. Lacking a clear physical intuition, Adam is difficult to generalize to manifolds. Some attempts have been made to directly apply parts of the Adam algorithm to manifolds or to find an underlying structure, but a full generalization has remained elusive.

In this work a new approach is presented that leverages the special structure of the manifolds which are relevant for optimization of neural networks, such as the Stiefel manifold, the symplectic Stiefel manifold and the Grassmann manifold: all of these are homogeneous spaces and as such admit a global tangent space representation. This is a common vector space, often called the Lie subspace, that makes the generalization of all steps in the Adam optimizer (as well as other optimizers) possible.

It is thus possible to extend the Adam optimizer to manifolds without a projection step, something that was not possible before. The resulting algorithm is then applied to train transformers and a symplectic autoencoder for which orthogonality constraints are enforced up to machine precision and we conclusively demonstrate the advantage of the proposed optimizer over existing methods.

\end{abstract}




\newpage

\tableofcontents

\newpage

\section*{Nomenclature}

\begin{tabular}{@{}ll}
    $\alpha$ & Momentum optimizer parameter \\
    $\beta_1, \beta_2$ & Exponential decay rates in Adam optimizer \\
    $\mathfrak{A}$ & Matrix function $\mathfrak{A}(X) = \sum_{k=1}^\infty X^{k-1}/k!$ used to evaluate $\exp$ on $\mathfrak{g}^{\mathrm{hor}}$ \\
    $\mathcal{B}$ & Part of the \textit{cache} (an element of $\mathfrak{g}^{\mathrm{hor},Y}$) \\
    $\Delta$ & Element of $T_Y\mathcal{M}$ (the Riemannian gradient in the algorithm) \\
    $D$ & Representation of $\Delta$ in $\mathfrak{g}^{\mathrm{hor}}$ \\
    $\delta$ & Parameter to avoid singularity in Adam optimizer \\
    $E$ & Distinguished element of a homogeneous space \\
    $\varepsilon$ & Machine precision \\
    $\exp$ & (Matrix) exponential map \\
    $\eta$ & Learning rate \\
    $\gamma$ & Geodesic \\
    $g$ & Riemannian metric \\
    $\mathfrak{g}$ & Lie algebra \\
    $\mathfrak{g}^{\mathrm{hor}}$ & Horizontal component of the Lie algebra with respect to $E\in\mathcal{M}$ \\
    $\mathfrak{g}^{\mathrm{hor}, Y}$ & Horizontal component of the Lie algebra with respect to $Y\in\mathcal{M}$ \\
    $\mathfrak{g}^{\mathrm{ver}, Y}$ & Vertical component of the Lie algebra with respect to $Y\in\mathcal{M}$ \\
    $G$ & Lie group \\
    $\mathbb{I}$ & Identity matrix \\
    $L$ & Loss function \\
    $\tilde{L}$ & Regularized loss function \\
    $\lambda$ & Section $Y\to{}G$ \\
    $\Lambda$ & Element in $G$ realizing a particular global section \\
    $\mathfrak{m}$ & Alternative notation for $\mathfrak{g}^{\mathrm{hor}}$ used in \cite{nomizu1954invariant,o1983semi,krogstad2003low,schlarb2024covariant,schlarb2024rolling} \\
    $\mathcal{M}$ & Riemannian manifold (homogeneous space in most cases) \\
    $\mu$ & Regularization parameter \\
    $\nabla{}L$ & Euclidean gradient of the loss function \\
    $\mathcal{O}$ & Optimizer \\
    $\mathbb{O}$ & Zero matrix \\
    $\Omega$ & Isomorphism that maps from $T_Y\mathcal{M}$ to $\mathfrak{g}^{\mathrm{hor},Y}$ \\
    $\pi$ & Projection from $G$ to $\mathcal{M}$ \\
    $\Pi$ & Parallel transport \\
    $\mathcal{R}$ & Retraction ($\mathcal{R}^{\mathrm{geo}}_Y$ the geodesic retraction) \\
    $\overline{\mathrm{retraction}}$ & Extended retraction (composition of \texttt{update\_section} and \texttt{apply\_section}) \\
    $\rho$ & Spectral radius \\
    $\sigma$ & Activation function of the feedforward layer ($\tanh$) \\
    $\Sigma$ & Regularizer \\
    $\mathfrak{so}(N)$ & Lie algebra of $SO(N)$ \\
    $SO(N)$ & Special orthogonal group \\
    $St(n, N)$ & Stiefel manifold \\
    $T_Y\mathcal{M}$ & Tangent space at $Y$ \\
    $t$ & Iteration index \\
    $\theta$ & Scaling threshold \\
    $\mathcal{V}$ & Vector space \\
    $V_Y$ & Element of $T_Y\mathcal{M}$ \\
    $W$ & Final velocity used to update the neural network parameters \\
    $\Xi$ & Collection of optimizer parameters \\
    $Y$ & Element of $\mathcal{M}$ (the neural network weights) \\
\end{tabular}

\newpage

\section{Introduction and Related Work}
\label{introduction}

The enormous success of neural networks in recent years has in large part been driven by the development of new and successful optimizers, most importantly Adam \cite{kingma2014adam}. Even though the Adam algorithm is nowadays used to train a wide variety of neural networks, its theoretical properties remain largely obscure \cite{kunstner2019limitations}. The implications of this are especially pertinent when attempting to generalize the Adam optimizer to the setting when neural network weights lie on manifolds.

Optimization on manifolds for training neural networks has emerged as an alternative to imposing soft constraints via extra terms in the loss function. Soft constraints modify the loss function in the following way:
\begin{equation}
    \tilde{L}(\mathrm{weights}; \mathrm{data}) = L(\mathrm{weights}; \mathrm{data}) + \mu\Sigma(\mathrm{weights}),
    \label{eq:regularized_loss}
\end{equation} where $\Sigma$ is called the ``regularizer'' \cite{goodfellow2016deep}. Two important classes of regularization terms are (i) orthogonality constraints (see \cite{zhang2021orthogonality} for an application to transformer neural networks and \cite{bansal2018can} for an application to convolutional neural networks) and (ii) properties relating to a physical system, yielding \textit{physics-informed neural networks} (PINNs, \cite{raissi2019physics}). Despite successful application in many different settings, an inherent problem is that these regularizers do not provide theoretical guarantees on the preservation of relevant properties and that they rely on extensive hyperparameter tuning. Regularization terms often make training neural networks difficult, cumbersome and ``unpredictable'' \cite{koch2018autotune}.

What would make these extra terms in the loss functional and the associated hyperparameters redundant is optimization on spaces that satisfy the associated constraints automatically \cite{becigneul2018riemannian, li2020efficient, kong2022momentum,lin2023simplifying,lezcano2019cheap}. Such spaces are in many cases \textit{homogeneous spaces} (see \cite[Chapter 11]{o1983semi} and \cite[Chapter 17]{frankel2011geometry}), a special class of manifolds. Homogeneous spaces that are important for neural networks include the Stiefel manifold \cite{li2020efficient, kong2022momentum}, the Grassmann manifold \cite{oja1989neural} and the ``space of positions and orientations'' \cite{van2024geodesic}. Recalling the two classes of regularization constraints above (i.e. (i) orthogonality constraints and (ii) physical constraints) we remark that for (i) we show in \Cref{transformer} that optimization on spaces that preserve orthogonality can enable training in some cases in which it otherwise would not be possible; in other cases it can speed up training immensely \cite{kong2022momentum}. Regarding point (ii), optimizing on manifolds can be inevitable for training specific neural networks\footnote{We show an example of such a system in \Cref{pendulum_sae} where we train a symplectic autoencoder \cite{brantner2023symplectic}.} \cite{brantner2023symplectic}. We sketch these two different motivations in \Cref{fig:manifold_optimization_motivation}.

\begin{figure}
    \centering
    \Huge 
\begin{tikzpicture}[y=0.80pt, x=0.80pt, yscale=-1.000000, xscale=1.000000, inner sep=0pt, outer sep=0pt,
                    motivation/.style={draw, very thick, rounded corners, minimum width=15ex},
                    physical/.style={motivation, fill=morange!30},
                    orthogonal/.style={motivation, fill=mblue!30},
                    arrow/.style={-{Latex[length=1cm, width=.5cm]}, ultra thick, rounded corners},
]

\node[physical, xshift = -6cm, yshift = -3cm] (phys_constraints) {Physical Constraints};
\node[below of=phys_constraints, orthogonal, yshift = -7.5cm] (orth_constraints) {Orthogonality Constraints};

\draw[fill=mpurple] (120.0000,420.0000) .. controls (71.8597,240.1403) and
  (39.4562,168.5088) .. (20.0000,180.0000) .. controls (216.5964,52.9473) and
  (434.1053,90.1579) .. (600.0000,300.0000) .. controls (472.8771,231.6315) and
  (325.8947,276.2807) .. (120.0000,420.0000) -- cycle;

\draw[fill=mgreen,opacity=0.500] (255.0000,140.0000) ellipse (0.4233cm and
  0.1693cm);

\draw[fill=mred] (50.0000,300.0000) .. controls (170.0000,153.3333) and
  (290.3860,98.8070) .. (403.7193,140.1403) .. controls (291.7193,101.4737) and
  (171.6667,158.0000) .. (51.0000,306.0000) -- (50.0000,300.0000) -- cycle;

\draw[fill=mpurple!50, draw opacity=0, fill opacity=0.5] (110.0000,300.0000) -- (20.0000,110.0000)
  -- (330.0000,70.0000) -- (610.0000,170.0000) -- (110.0000,300.0000);

\draw[fill=mred!50, fill opacity=0.516, draw opacity=0, line width=0.800pt, cm={{0.94667,-0.32222,0.32222,0.94667,(0.0,0.0)}}, miter limit=4.00]
  (45.3210,210.4979) -- (198.7844,211.7622) .. controls (200.5020,212.3468) and
  (201.1998,215.5767) .. (198.7844,216.4794) -- (45.6572,216.4794) --
  (45.3210,210.4979) -- cycle;

\draw[fill=mred] (110.7360,184.6825) -- (256.4219,136.4306) .. controls (258.2362,136.4306)
  and (259.0531,138.2428) .. (258.6902,139.8983) -- (111.5302,189.5110) --
  (110.7360,184.6825) -- cycle;

\draw[fill=mred!50, fill opacity=0.516, draw opacity=0] (90.0000,195.0000) -- (114.6371,176.4514) .. controls (121.6667,174.6962)
  and (127.1772,189.5230) .. (121.7343,194.0800) -- (90.0000,195.0000) -- cycle;

\draw[fill=mred] (90.0057,195.0148) -- (114.6429,176.4663) .. controls (121.6724,174.7110)
  and (124.9604,185.0928) .. (123.5543,191.1013) -- (90.0057,195.0148) -- cycle;

\draw[fill=black] (6.0000,321.0000)   node[above right] () {$c (t)$};
\draw[fill=black] (69.0000,186.0000)  node[above right] () {$v$};
\draw[fill=black] (272.0000,167.0000) node[above right] () {$q$};
\draw[fill=black] (316.0000,49.0000)  node[above right, yshift=-.5cm] () {${T}_{q} \mathcal{M}$};
\draw[fill=black] (419.0000,318.0000) node[above right] () {$\mathcal{M}$};

\draw[arrow, black!30!morange] (phys_constraints.east) to [bend right = 15] node[midway, above, yshift=.2cm] {\cite{brantner2023symplectic}} (245.0000,133.0000);
\draw[arrow, black!30!mblue] (orth_constraints.east) to [bend left = 15] node[midway, below, rotate=30, xshift=-2.98cm, yshift=-.4cm] {\cite{lezcano2019cheap,li2020efficient,kong2022momentum}} (245.0000,153.0000);

\end{tikzpicture}
    \caption{The motivation for manifold optimization comes from different angles: (i) for many applications manifold optimization can bring huge improvements in training the network through e.g. automatically enforced orthogonality constraints \cite{lezcano2019cheap,li2020efficient,kong2022momentum}; (ii) for applications in physics certain network architectures, like the ones presented in \cite{brantner2023symplectic}, require manifold optimization. The transformer and symplectic-autoencoder examples in this work illustrate these two motivations.}
    \label{fig:manifold_optimization_motivation}
\end{figure}

In order to fully benefit from introducing manifolds into neural networks, existing powerful optimizers should be extended to this setting, but the obscure nature of the Adam algorithm has meant that its properties have to be severely restricted before this extension can be made.\footnote{This is not true for optimizers that only contain first moments, but no second moments (like the Adam optimizer). Optimizers that only contain first moments admit a variational formulation and allow for straightforward generalization to arbitrary manifolds \cite{duruisseaux2022accelerated, kong2022momentum}.} We note that this is a well-known problem \cite{becigneul2018riemannian, lezcano2019cheap}. In \cite{lezcano2019cheap} the authors say: ``there is a vast literature on optimization methods on Riemannian manifolds [...]. On the other hand, the problem of adapting popular optimization algorithms like [Adam] is a topic of current research.'' The problem with Adam is that it cannot, unlike other Euclidean optimization methods such as the Newton method, be transferred to the general manifold setting as outlined in e.g. \cite{absil2008optimization,boumal2023intromanifolds}. Mathematically the Adam optimizer depends on the choice of a specific coordinate system, or put differently \cite{becigneul2018riemannian}: ``the adaptivity of these algorithms [i.e. Adam] can be thought of as assigning one learning rate per coordinate of the parameter vector. However, on a Riemannian manifold, one is generally not given an intrinsic coordinate system, rendering meaningless the notions sparsity or coordinate-wise update.''

In this work we pick a \textit{global tangent space representation} in order to generalize Adam to homogeneous spaces. For Lie groups, a special case of homogeneous spaces, this global tangent space representation is the Lie algebra $\mathfrak{g}$. For Riemannian homogeneous spaces this \textit{global tangent space representation} is the horizontal component of the associated Lie algebra at a distinct element $E$. This horizontal component is denoted by $\mathfrak{m}$ in \cite{nomizu1954invariant,o1983semi,krogstad2003low,schlarb2024covariant,schlarb2024rolling} and called a ``Lie subspace'' in \cite{o1983semi}, and denoted by $\mathfrak{g}^\mathrm{hor}$ in this work. The use of $\mathfrak{g}^\mathrm{hor}$ (respectively $\mathfrak{m}$) in computations involving the Stiefel manifold was previously discussed by e.g. \cite[Section 2.3.1]{edelman1998geometry}.

Viewing $\mathfrak{g}$ as the global tangent representation for Lie groups was leveraged in \cite{lezcano2019cheap} as well as \cite[Algorithm 5]{kong2022momentum} to generalize the Adam optimizer to Lie groups. As the authors in \cite{lezcano2019cheap} write, one would have to extend their results to ``homogeneous Riemannian manifolds, like the Stiefel manifold'' to generalize the Adam optimizer to more general neural network architectures. In this work we aim to make this extension.

We first state the contribution of this work and then give an outline of the structure of the remainder of this text.

\subsection{Our Contribution}

In this work we present a generalization of the Adam algorithm to homogeneous spaces such as the Stiefel manifold. Algorithms with the same goal were already proposed in \cite[Algorithm 2]{li2020efficient} and \cite[Algorithm 2]{kong2022momentum}, but these algorithms modify Adam in such a way that the original idea of coordinate-wise updates is lost; it has to be noted that for Lie groups (i.e. the case when $\mathfrak{g}=\mathfrak{g}^\mathrm{hor}$), the approach presented in this paper reduces to \cite[Algorithm 5]{kong2022momentum} or can be derived with the theory outlined in \cite{lezcano2019cheap}. To state precisely what is new and what is inherited, we separate the contribution into distinct parts:

\begin{itemize}
    \item \textbf{The construction.} A global horizontal tangent-space representation in which coordinate-wise Adam updates can be applied on a homogeneous space: in each step of the iteration an element of the tangent space $T_Y\mathcal{M}$ is converted to an element of $\mathfrak{g}^\mathrm{hor}$, and in this space all of the steps necessary for the Adam optimizer are performed.
    \item \textbf{The scope.} The construction extends the Lie-group case (i.e. \cite{lezcano2019cheap} and \cite[Algorithm 5]{kong2022momentum}) to homogeneous spaces such as the Stiefel and the Grassmann manifold.
    \item \textbf{The evidence.} Controlled comparisons on the vision-transformer task of \Cref{transformer} as well as the symplectic model reduction task of \Cref{pendulum_sae}, in which all optimizer configurations share the architecture and hyperparameters, and for which the cost and runtime overhead are reported. We thus conclusively demonstrate the effectiveness of the new optimizer(s).
\end{itemize}

We further want to stress that the idea of representing an element of $T_Y\mathcal{M}$ in $\mathfrak{g}^\mathrm{hor}$ itself is not novel as it already appears in e.g. \cite{edelman1998geometry,bendokat2021real}, but the idea of utilizing this representation to generalize neural network optimizers is new. We thus manage, as is stated in \cite{kong2022momentum} for the Lie group $SO(N)$, that ``the momentum will always stay in the same tangent space'' and this ``almost reproduce [sic] the convenience in Euclidean space for $SO(N).$''

We summarize the comparison of different methods below:

\noindent\begin{tabular}{lcccc}
    \toprule
    & on $SO(N)$ & on $St(n, N)$ & vec.-valued $\mathcal{B}^\mathtt{cache}_2$ & no projections  \\
    \midrule
    \cite{lezcano2019cheap} & \cmark & \xmark & \cmark & \cmark \\
    \cite[Algorithm 5]{kong2022momentum} & \cmark & \xmark & \cmark & \cmark \\
    \addlinespace
    \cite[Algorithm 2]{li2020efficient} & \cmark & \cmark & \xmark & {\color{gray}not applicable} \\
    \cite[Algorithm 2]{kong2022momentum} & \cmark & \cmark & \cmark & \xmark \\
    \midrule
    Proposed method & \cmark & \cmark & \cmark & \cmark \\
    \bottomrule
\end{tabular}

\subsection{Outline}

The remainder of this paper is organized as follows: \Cref{homogeneous_spaces} gives a description of all the relevant aspects of homogeneous spaces that are needed to extend the Adam optimizer the way it is done here. This does not include original work and summarizes relevant parts of \cite{o1983semi,edelman1998geometry}.

In \Cref{adam_section} we introduce the Adam optimizer and explain the difficulties in generalizing it to manifolds. This section also contains a review of related approaches and their shortcomings (see \Cref{existing_approaches}).

Building on our material from previous sections, the generalization of Adam to homogeneous spaces is described in \Cref{general_framework}; in \Cref{transformer} and \Cref{pendulum_sae} we apply it to a transformer and a symplectic autoencoder, respectively. The new optimizers are part of the \texttt{Julia} package \texttt{GeometricOptimizers.jl} \cite{brantner2026geometricoptimizers}.

Throughout the discussion, the Stiefel manifold will be used as an example to elucidate the more abstract and general concepts; however the mathematical constructions are more general and can, in addition to the Stiefel manifold, be applied to the Grassmann manifold \cite{bendokat2020grassmann}, the symplectic versions of the two manifolds \cite{bendokat2021real} as well as the ``homogeneous space of positions and orientations'' \cite{van2024geodesic}. The way we generalize Adam here is also not limited to that optimizer, but can also be applied to all other (first-order) optimization methods like RMSProp, AdaGrad \cite{duchi2011adaptive} or BFGS \cite{wright2006numerical}.

\section{Homogeneous Spaces and a Global Tangent Space Representation}
\label{homogeneous_spaces}

In this section we discuss some aspects of homogeneous spaces that are relevant for our algorithm. The contents of this section are largely taken from \cite[Chapter 11]{o1983semi} and \cite[Section 2]{edelman1998geometry}. The Stiefel manifold is used as a concrete example to make abstract constructions more clear, but what is presented here also applies to all homogeneous spaces such as the Grassmann manifold \cite{bendokat2020grassmann} and the ``homogeneous space of positions and orientations'' \cite{van2024geodesic}. A central object in this section is a \textit{global tangent space representation} that simply is the Lie algebra $\mathfrak{g}$ when dealing with a Lie group and a ``Lie subspace'' \cite{o1983semi} when dealing with a homogeneous space. We will denote such a space by $\mathfrak{g}^\mathrm{hor}$. In \cite{nomizu1954invariant,o1983semi,krogstad2003low,schlarb2024covariant,schlarb2024rolling} this is denoted by $\mathfrak{m}$.

\begin{definition}\label{def:homogeneous_spaces}
A \textbf{homogeneous space} is a manifold $\mathcal{M}$ on which a Lie group $G$ acts transitively, i.e. the group action $G\times\mathcal{M}\to\mathcal{M}$ is such that $\forall{}Y_1,Y_2\in\mathcal{M}\quad\exists{}\Lambda\in{}G$ s.t. $l_\Lambda{}Y_1 \equiv \Lambda{}Y_1 =  Y_2$ (here $l_\Lambda$ denotes the left action of an element $\Lambda$ of $G$ onto $\mathcal{M}$).
\end{definition}

In the following we will associate a distinct element\footnote{In the case when the homogeneous space is a vector space $\mathcal{V}$ then the group action is addition and the distinct element is the zero element: $E = \mathbb{O}\in\mathcal{V}$.} $E_\mathcal{M} \equiv E$ with every homogeneous space $\mathcal{M}$. Because $G$ acts transitively on $\mathcal{M}$, every element of $\mathcal{M}$ can then be represented with an element of $G$ acting on $E$. We further note that in \Cref{def:homogeneous_spaces} we used the notation $\Lambda\in{}G$ in anticipation of $\Lambda$ usually being the result of applying a section $\lambda:\mathcal{M}\to{}G$ (see \Cref{def:section}).
As an example of a homogeneous space, consider the orthogonal group $SO(N) := \{\lambda\in\mathbb{R}^{N\times{}N}: \lambda^T\lambda = \mathbb{I}_N\}$ and the Stiefel manifold $\mathcal{M}_\mathrm{Stiefel} \equiv St(n, N) := \{Y\in\mathbb{R}^{N\times{}n}: Y^TY = \mathbb{I}_n\}$ where $n\leq{}N$. An element of $SO(N)$ is a collection of $N$ orthonormal vectors in $\mathbb{R}^N$ and an element of $\mathcal{M}_\mathrm{Stiefel}$ is a collection of $n$  orthonormal vectors in $\mathbb{R}^N$. As distinct element\footnote{We note that this distinct element $E$ was already given a distinct role in \cite{edelman1998geometry}, in whose terminology $E$ would be called $I_{N,n}$ and is referred to as ``the origin''. The same terminology also appears in \cite{bendokat2021real}.} for $\mathcal{M}_\mathrm{Stiefel}$ we pick:

\begin{equation}
    E = \begin{bmatrix}
        1 & 0 &\ldots & 0 \\
        0 & 1 & \ldots & 0 \\
         & \ldots & \ldots & \\
         0 & 0 & \ldots & 1 \\
         0 & 0 & \ldots & 0 \\
         & \ldots & \ldots & \\
         0 & 0 & \ldots & 0
    \end{bmatrix} = \begin{bmatrix} \mathbb{I}_n \\ \mathbb{O}  \end{bmatrix}\text{, where }\mathbb{O}\in\mathbb{R}^{(N-n)\times{}n}.
\end{equation}

If we now apply an element $[y_1, \ldots,y_n, y_{n+1}, \ldots,  y_N] = \Lambda\in{}G$ from the left onto $E$, the resulting matrix consists of the first $n$ columns of $\Lambda$: $\Lambda{}E = Y = [y_1, \ldots, y_n]$. So multiplication with $E$ from the right maps $G$ to $\mathcal{M}$. We will also write $\pi(\Lambda) := \Lambda{}E$. With this we can define the notion of a \textit{section}:

\begin{definition}\label{def:section}
    A \textbf{section} is a mapping $\lambda:\mathcal{M}\to{}G$ s.t. $\pi(\lambda(Y)) = \lambda(Y)E = Y$.
\end{definition}

Sections are necessary for the computation of a map from the tangent space $T_Y\mathcal{M}$ to the global tangent space representation\footnote{We call the corresponding map \texttt{global\_rep} (see \Cref{op:global_rep}).} $\mathfrak{g}^\mathrm{hor}$. We now give a concrete example of a section:
Consider the element $Y = [y_1, y_2, \ldots, y_n]\in\mathcal{M}$. We want to map it to its associated Lie group $G = SO(N)$. In order to do so we have to find $(N-n)$ orthonormal vectors that are also orthonormal to $Y$; i.e. we have to find $Q = [q_1, \ldots, q_{N-n}]$ such that $Y^TQ = \mathbb{O}_n$ and $Q^TQ = \mathbb{I}_n$.
In the implementation this is done using a $QR$ decomposition (Householder reflections). Details of this are shown in \Cref{alg:lift} and further discussed in \Cref{global_section}.

\begin{algorithm}
    \caption{Computation of the lift $Y\mapsto\lambda(Y)\in{}G$ with a $QR$ decomposition.}\label{alg:lift}
    \begin{algorithmic}
    \State $A\gets{}\mathcal{P}(\mathbb{R}^{N\times(N-n)})$, \Comment{sample $A$ from a given distribution.}
    \State $A \gets A - YY^TA$, \Comment{remove the part of $A$ that is spanned by the columns of $Y$.}
    \State $Q,R\gets\mathtt{qr}(A)$, \Comment{apply a $QR$ decomposition.}
    \State $\lambda(Y)\gets [Y, Q\mathtt{[1:N,1:(N-n)]}]$. \Comment{output $Y$ and the first $(N-n)$ columns of $Q$.}
    \end{algorithmic}
\end{algorithm}

\begin{remark}
    We note that this section is non-unique as it depends on the sampled $A$. In \Cref{alg:general_man} we however only have to perform this operation once when we initialize the optimizer. For all successive steps we \textit{parallel transport} (see \Cref{parallel_transport}) the section $\Lambda^{(t)}$ with \texttt{update\_section} (see \Cref{alg:general_man} and \Cref{fig:new_optimizer}).
\end{remark}

We now proceed with introducing the remaining necessary components that are needed to define the global tangent space representation $\mathfrak{g}^\mathrm{hor}$ (this is referred to as a \textit{Lie subspace} in \cite{o1983semi}). In the following we assume that the Lie group $G$ is equipped with a Riemannian metric\footnote{As a metric $g$ on $G=SO(N)$ we take the canonical one, i.e. $g:(V_1,V_2) \mapsto \frac{1}{2}\mathrm{Tr}(V_1,V_2)$.} $g$.

First note that $T_Y\mathcal{M} \equiv \frak{g}\cdot{}Y$, i.e. the tangent space at $Y$ can be represented by the application of the Lie algebra of $G$ to the element $Y$. The kernel of this map $\frak{g}^{\mathrm{ver},Y}:=\mathrm{ker}(\frak{g}\to{}T_Y\mathcal{M})$ is called the \textit{vertical component} of the Lie algebra at $Y$.
Its complement (with respect to the Riemannian metric $g$ on $G$) in $\mathfrak{g}$ is called the \textit{horizontal component} of $\mathfrak{g}$ at $Y$ and denoted by $\mathfrak{g}^{\mathrm{hor},Y}$; this therefore establishes an isomorphism $\mathfrak{g}^{\mathrm{hor},Y}\simeq{}T_Y\mathcal{M}$ and the mapping $\Omega:T_Y\mathcal{M}\to\mathfrak{g}^{\mathrm{hor},Y}$ can be found explicitly\footnote{This mapping was derived before in e.g. \cite[Section 3.]{celledoni2003implementation} and denoted by $a_Q$ for $Q\in{}St(n,N)$. In \cite{krogstad2003low,schlarb2024rolling} the space $\mathfrak{g}^{\mathrm{hor}, Y}$ is called $\mathfrak{m}_Y$.}. For the Stiefel manifold this mapping is (in \Cref{induced_isomorphism} we show that $\Omega$ is indeed an isomorphism):

\begin{equation}
    \Omega(V_Y) = \bigg(\mathbb{I} - \frac{1}{2}YY^T\bigg)(V_Y)Y^T - Y(V_Y)^T\bigg(\mathbb{I} - \frac{1}{2}YY^T\bigg),
    \label{eq:omega}
\end{equation}

where $V_Y\in{}T_Y\mathcal{M}$. Later we will use $\Omega$ to map the Riemannian gradient (see \Cref{op:rgrad}) $\mathrm{grad}_YL\in{}T_Y\mathcal{M}$ to $\mathfrak{g}^\mathrm{hor}$. Also confer \Cref{fig:new_optimizer} to see where this mapping appears in our algorithm. The mapping $\Omega$ naturally induces a metric on the Stiefel manifold: two vectors of $T_Y\mathcal{M}$ can be mapped to $\mathfrak{g}^{\mathrm{hor},Y} \subset\mathfrak{g}$ and the scalar product can be computed in the Lie algebra, i.e. $\Omega$ established an isometry between $T_Y\mathcal{M}$ and $\mathfrak{g}^{\mathrm{hor},Y}$ \cite[Chapter 11]{o1983semi}. If $G = SO(N)$ is equipped with the canonical metric one obtains the canonical metric for the Stiefel manifold shown in \Cref{eq:stiefel_manifold_metric}. This equation was previously derived in e.g. \cite[Equation (2.39)]{edelman1998geometry}. We can now define what we mean by \textit{global tangent space representation for a homogeneous space}:

\begin{definition}\label{def:global_tangent_space}
    To each homogeneous space we can associate a \textbf{global tangent space representation} $\mathfrak{g}^{\mathrm{hor},E}$ where we have identified a distinct element $E\in\mathcal{M}$. It is the horizontal component of the Lie algebra at the distinct element $E$. We typically simply call this $\mathfrak{g}^\mathrm{hor}$.
\end{definition}

\begin{remark}
    Of course every manifold has such a global tangent space representation, because $T_Y\mathcal{M}\simeq\mathbb{R}^d$ where $d=\mathrm{dim}\mathcal{M}$. But usually the mappings $T_Y\mathcal{M}\to\mathbb{R}^d$ are very difficult to find. What we exploit here is that for homogeneous spaces this is relatively easy.
\end{remark}

Mapping an element of $T_Y\mathcal{M}$ to the global tangent space representation $\mathfrak{g}^\mathrm{hor}$ requires another mapping\footnote{In this text we summarize the composition of those two mappings, i.e. $\Omega:T_Y\mathcal{M}\to\mathfrak{g}^{\mathrm{hor},Y}$ and $\mathfrak{g}^{\mathrm{hor},Y}\to\mathfrak{g}$, under the name \texttt{global\_rep} (see \Cref{op:global_rep} and \Cref{fig:new_optimizer}).} besides $\Omega$; one from  $\mathfrak{g}^{\mathrm{hor},Y}$ to $\mathfrak{g}^{\mathrm{hor}}\equiv\mathfrak{g}^{\mathrm{hor},E}$.

We obtain this second mapping by first computing\footnote{In practice computing a section is only done when initializing the optimizer. For this we use \Cref{alg:lift}.} or updating\footnote{For updating the section (i.e. for successive steps after the optimizer has been initialized) we use parallel transport (see \Cref{parallel_transport}).} a \textit{section} of $Y$ (see \Cref{def:section}) and then performing:
\begin{equation}
    \Omega(V_Y)\mapsto \lambda(Y)^{-1}\Omega(V_Y)\lambda(Y).
    \label{eq:ghorYghor}
\end{equation}

The following theorem shows that equation \Cref{eq:ghorYghor} establishes an isomorphism:

\begin{theorem}
    The map $\mathfrak{g}^{\mathrm{hor},Y}\to\mathfrak{g}^{\mathrm{hor}}$, $Z\mapsto\Lambda^{-1}Z\Lambda$ for any $\Lambda\in{}G$ such that $\Lambda{}E = Y$ is an isomorphism.
\end{theorem}
\begin{proof}
    The considered map is clearly invertible. We still have to show that $\Lambda^{-1}Z\Lambda$ is an element of $\mathfrak{g}^{\mathrm{hor}}$ and that for every element $B\in\mathfrak{g}^\mathrm{hor}$ there exists $Z_B\in\mathfrak{g}^{\mathrm{hor},Y}$ such that $\Lambda^{-1}Z_B\Lambda = B$.

    For the first part, assume $\Lambda^{-1}Z\Lambda\notin\mathfrak{g}^{\mathrm{hor}}$. This implies that there exists a decomposition $\Lambda^{-1}Z\Lambda = V_Z + H_Z$ with $V_Z\neq 0$ being the vertical part, i.e. $V_ZE = 0$. But then $Z = \Lambda{}V_Z{}\Lambda^{-1} + \Lambda{}H_Z\Lambda^{-1}$  with $0 \neq\Lambda{}V_Z{}\Lambda^{-1}\in\mathfrak{g}^{\mathrm{ver},Y}$. A contradiction.

    For the second part, consider $B\in\mathfrak{g}^\mathrm{hor}$. By a similar argument as above we can show that $\Lambda{}B\Lambda^{-1}\in\mathfrak{g}^{\mathrm{hor},Y}$. This element fulfills our requirements.
\end{proof}

We further note that the particular space $\mathfrak{g}^{\mathrm{hor}}\equiv \mathfrak{g}^{\mathrm{hor},E}$, the horizontal subspace of $\mathfrak{g}$ at $E$, allows for an especially sparse and convenient representation. For the Stiefel manifold this is shown in e.g. \cite[Equation (2.21)]{edelman1998geometry}. We recall it here:
\begin{equation}
    \mathfrak{g}^\mathrm{hor} = \left\{
            \begin{pNiceMatrix}[margin=.5em]
            A       &      \text{ } & -B^T & \text{ } \\
            \text{ }&      \text{ } &      &       \\
            B       &      \text{ } &  0   &       \\
            \text{ }&      \text{ } &      &
            \CodeAfter
                \SubMatrix[{1-2}{1-4}]
                \SubMatrix[{2-1}{4-1}]
                \UnderBrace[right-shorten,yshift=3pt]{1-1}{4-1}{n}
                \UnderBrace[left-shorten,yshift=3pt]{1-2}{4-4}{N-n}
            \end{pNiceMatrix}: A\in\mathbb{R}^{n\times{}n}\text{ skew-sym, } B\in\mathbb{R}^{N\times(N-n)}\text{ arbitrary}
           \right\}.
  \label{eq:g_hor}
\end{equation}

\vspace{.5cm}

\subsection{Retractions on Homogeneous Spaces}

In order to perform optimization on manifolds the concept of a \textit{retraction} \cite{absil2008optimization, boumal2023intromanifolds} is essential. This in turn is an approximation of a geodesic. We now give definitions of these concepts for general Riemannian manifolds and then recall a specific statement regarding homogeneous spaces in \Cref{th:geodesic}.

\begin{definition}\label{def:geodesic}
    A \textbf{geodesic} is the solution of the geodesic differential equation. It is the curve that minimizes the following functional:
    \begin{equation}
        L(\gamma) = \int_0^1\sqrt{g_{\gamma(t)}(\dot{\gamma}(t), \dot{\gamma}(t))} \text{ s.t. $\gamma(0) = a$ and $\gamma(1) = b$ are fixed endpoints}.
    \end{equation}
\end{definition}

So a geodesic is a curve that minimizes the distance between the points $a$ and $b$. Closely connected to the notion of a geodesic is that of the \textit{geodesic spray}. The \textit{geodesic spray} or \textit{geodesic differential equation} \cite[chapter 5]{bishop1980tensor} is a second order ordinary differential equation on the manifold $\mathcal{M}$, so takes initial conditions on $T_Y\mathcal{M}$. Its solutions have the following property\footnote{Here we use $T\mathcal{M}$ to denote the \textit{tangent bundle}. This is the union of all tangent spaces, i.e. $T\mathcal{M} = \cup_{Y\in\mathcal{M}}T_Y\mathcal{M}$, and can be given the structure of a $2N$-dimensional manifold \cite{bishop1980tensor}.}:
\begin{proposition}
    Let $\bar{\gamma}:[0, T]\to{}T\mathcal{M}$ be a solution to the geodesic spray. It has the property that
    \begin{equation}
        \bar{\gamma}(t) = (\gamma(t), \gamma'(t)),
    \end{equation}
    and $\gamma$ is a geodesic.
\end{proposition}

For Riemmanian optimization problems \cite{absil2008optimization, sato2021riemannian} this differential equation has to be solved exactly or approximately. If it is solved approximately\footnote{We further note that even though we use retractions for approximating geodesics (see \Cref{def:geodesic}) here  (and this is described as their ``key property'' in \cite{absil2012projection}), retractions can be defined independently of geodesics and do not even require a Riemannian metric.}, we refer to the method of approximating it at as a \textit{classical retraction} \cite[Definition 3.47]{boumal2023intromanifolds}. We recall its definition here:

\begin{definition}\label{def:classical_retraction}
    A \textbf{classical retraction} is a smooth map $\mathcal{R}:T\mathcal{M} \to \mathcal{M}$ that satisfies the following properties: $\mathcal{R}_Y(\mathbb{O}_Y) = Y$ and $T_{\mathbb{O}_Y}\mathcal{R}_Y = \mathrm{id}|_{T_Y\mathcal{M}}$ where $\mathbb{O}_Y$ is the zero element of the vector space $T_Y\mathcal{M}$ and $\mathrm{id}|_{T_Y\mathcal{M}}$ is the identity map on $T_Y\mathcal{M}$.
\end{definition}

We note that we use \textit{extended retractions} in our implementation of the proposed algorithm (see \Cref{alg:general_man}) instead of \textit{classical retractions}. \textit{Extended retractions} are defined in \Cref{op:retraction}. We recall a theorem that serves a basis for reducing the search for retractions on homogeneous spaces to the search for retractions on Lie groups:

\begin{theorem}
    Let $\Delta$ be an element of the tangent space $T_Y\mathcal{M}$ of a Riemannian homogeneous space $\mathcal{M}$ with Lie group $G$ and metric $g$. Then the associated geodesics are $\gamma_\mathcal{M}(t;\Delta) = \gamma_G(t;\Omega(\Delta))Y$, where $\gamma_G$ denotes the geodesic on $G$ and $\Omega:T_Y\mathcal{M}\to\mathfrak{g}^{\mathrm{hor},Y}$ is the horizontal lift that was defined above.
    \label{th:geodesic}
\end{theorem}
\begin{proof}
    See \cite[corollary 7.46]{o1983semi}.
\end{proof}

\Cref{th:geodesic} is useful as for $G = SO(N)$ the geodesic map $\mathfrak{g}\to{}G$ is just the matrix exponential map \cite[Example 5.4.2]{absil2008optimization}. This means that for the Stiefel manifold the geodesic for $\Delta\in{}T_Y\mathcal{M}$ takes the following form:
\begin{equation}
    \gamma_\mathcal{M}(t;\Delta) = \gamma_G(t;\Omega(\Delta))Y = \exp(t\Omega(\Delta))\lambda(Y)E = \lambda(Y)\exp(t\mathcal{B})E,
    \label{eq:geodesic}
\end{equation}
where $\mathcal{B}$ is the representation of $\Omega(\Delta)$ in $\mathfrak{g}^\mathrm{hor}$, i.e. $\mathcal{B} = \lambda(Y)^{-1}\Omega(\Delta)\lambda(Y)$ for a global section $\lambda(Y)$. It has also been used that:
\begin{equation}
    \begin{aligned}
    \exp(t\Omega(\Delta)) = \exp(t\lambda(Y)\mathcal{B}\lambda(Y)^{-1}) & = \sum_{n=0}^\infty\frac{1}{n!}(t\lambda(Y)\mathcal{B}\lambda(Y)^{-1})^n \\ & = \lambda(Y)\left(\sum_{n=0}^\infty \frac{1}{n!}(t\mathcal{B})^n\right)\lambda(Y)^{-1}.
    \end{aligned}
\end{equation}
Because of the specific structure of $\mathcal{B}$ (see \Cref{eq:g_hor}) the computation of the exponential is very cheap \cite{bendokat2021real, celledoni2000approximating}. This is further discussed in \Cref{computing_the_exponential}. We also note that even though the global section $\lambda(Y)$ is not unique, the geodesic in \Cref{eq:geodesic} is, independent of the particular choice of $\lambda(Y)$. Put differently the ambiguity in the choice of $\lambda(Y)$ corresponds to the multiplication of $Y$ by an element of the ``isotropy group'' \cite[Section 2]{krogstad2003low} (also see \cite{munthe2016integrators}). We further note that the geodesic on $G$ defined through:
\begin{equation}
    g(t) = \lambda(Y)\exp(t\mathcal{B}),
\end{equation}
is a horizontal geodesic (see \cite[Definition 7.44]{o1983semi} and below), meaning that $g'(t)g(t)^{-1}\in\mathfrak{g}^{\mathrm{hor},\pi(g(t))}$ or written differently $\Omega\circ{}T\pi(g'(t)) = g'(t)\exp(-t\mathcal{B})\lambda(Y)^{-1}$.

\section{The Adam Optimizer}
\label{adam_section}

The Adam optimizer, along with RMSProp and AdaGrad \cite{duchi2011adaptive}, belongs to a family of first-order optimizers that update a cache in every optimization step based on non-trivial operations on the gradient of the loss function. Based on this cache these algorithms then compute a \textit{final velocity} with which the neural network weights are updated. The idea is shown in \Cref{alg:adam_alg} for the case when all weights lie on a Euclidean vector space.

\begin{algorithm}
    \caption{High-level optimization algorithm for neural networks}\label{alg:adam_alg}
    \begin{algorithmic}
    \Require time $t$, weights $Y^{(t)}$, \texttt{cache}, $\mathcal{O}$-parameters $\Xi$;
    \State $t \gets t + 1$
    \State $\Delta^{(t)} \gets \nabla_{Y^{(t)}}L$ \Comment{compute the Euclidean gradient.}
    \State $\mathtt{cache} \gets \mathtt{update}(\mathtt{cache}, \Delta^{(t)}, t, \Xi)$ \Comment{update cache based on $\Delta^{(t)}$.}
    \State $W^{(t)} \gets \mathtt{velocity}(\mathtt{cache}, \Xi)$ \Comment{compute \textit{final velocity}.}
    \State $Y^{(t+1)} \gets Y^{(t)} + W^{(t)}$ \Comment{add \textit{final velocity} to $Y^{(t)}$.}
    \end{algorithmic}
\end{algorithm}

\Cref{alg:adam_alg} defines a high-level optimizer method:

\begin{definition}[optimizer method]
    The \texttt{update!} method, \texttt{velocity} method and the \texttt{cache} together constitute an \textbf{optimizer method}. The \texttt{update!} method updates the cache and the \texttt{velocity} method computes a final velocity. This is then used to update the neural network parameters.
    \label{dftn:method}
\end{definition}

For the Adam optimizer the \textit{cache} consists of \textit{first} and \textit{second moments}: $\mathtt{cache} = (\mathcal{B}_1^\mathtt{cache}, \mathcal{B}_2^\mathtt{cache})$. Both of these are initialized with zeros and updated with:

\begin{equation}
    \begin{aligned}
    \mathcal{B}_1^\mathtt{cache} \gets &  \frac{\beta_1 - \beta_1^t}{1 - \beta_1^t}\mathcal{B}_1^\mathtt{cache} + \frac{1 - \beta_1}{1 - \beta_1^t}\mathcal{B}^{(t)},\\
    \mathcal{B}_2^\mathtt{cache} \gets &  \frac{\beta_2 - \beta_2^t}{1 - \beta_2^t}\mathcal{B}_2^\mathtt{cache} + \frac{1 - \beta_2}{1 - \beta_2^t}\mathcal{B}^{(t)}\odot{}\mathcal{B}^{(t)},
    \end{aligned}
    \label{eq:adam_cache}
\end{equation}

where $\odot$ is the elementwise Hadamard product, i.e. $(v\odot{}w)_i := v_iw_i$. In \Cref{alg:adam_alg} we summarize these two steps as \texttt{update} and the parameters $\beta_1$ and $\beta_2$ are stored in $\Xi$. After having updated the cache, a \textit{final velocity} is computed. For Adam:

\begin{equation}
    W^{(t)}\gets \mathtt{velocity}(\mathtt{cache}, \Xi) = -\eta\mathcal{B}^\mathtt{cache}_1/\sqrt{\mathcal{B}^\mathtt{cache}_2 + \delta},
    \label{eq:adam_velocity}
\end{equation}
where the \textit{learning rate} $\eta$ is again stored in $\Xi$.

\begin{remark}
    Modern optimization of neural networks is almost always done with some version of gradient descent that takes two inputs: a differentiated loss function $\nabla_YL\equiv\nabla{}L$, which is the output of an automatic differentiation (AD)\footnote{AD underpins the training of practically every deep neural network, but its discussion would be beyond the scope of this paper; \cite{griewank2003mathematical, bolte2020mathematical} offer a rigorous and detailed discussion of AD. For the purposes of this paper an AD routine simply takes a loss function $L$ as input and returns its (Euclidean) gradient with respect to the weights of the neural network $\nabla_Y{}L \equiv \nabla{}L$.} routine, and a \texttt{cache} that stores information about previous gradient descent steps (see e.g. \cite[chapter 8]{goodfellow2016deep}). Adam is one example of such an optimizer, i.e. a variation of gradient descent that stores second moments in the \texttt{cache}.
\end{remark}

\subsection{Existing Approaches towards Generalizing Adam to Manifolds}
\label{existing_approaches}

Various approaches have been proposed to generalize \Cref{alg:adam_alg} to homogeneous spaces or even more general manifolds. The main problem with this generalization is that the operations like the Hadamard product $\odot$ and the elementwise division are coordinate-dependent and are therefore not easily generalizable to arbitrary manifolds.

A first, partly successful, attempt to get around this problem was proposed in \cite{li2020efficient}. In there the authors approximate $\mathcal{B}_2^\mathtt{cache}$ with a scalar quantity, i.e.

\begin{equation}
    \mathcal{B}^\mathtt{cache}_2  \gets  \frac{\beta_2 - \beta_2^t}{1 - \beta_2^t}\mathcal{B}^\mathtt{cache}_2 + \frac{1 - \beta_2}{1 - \beta_2^t}||\mathcal{B}^{(t)}||^2,
    \label{eq:scalar_adam}
\end{equation}

which leads to speed ups in some cases, but ignores the structure of the second moments. The approach presented in \cite{kong2022momentum} is more involved and tries to address this problem\footnote{We also refer to \cite{kong2022momentum} for the excellent overview of existing approaches, especially regarding different methods for constructing retractions (see \Cref{def:classical_retraction}).}. The authors formulate the optimization problem as an unconstrained variational problem with Lagrangian on the Stiefel manifold:

\begin{equation}
    \mathfrak{L}(Y,\dot{Y}, \bar{A}, t) = r(t)\left[\frac{1}{2}\mathrm{Tr}\left(\dot{Y}^T(\mathbb{I} - aYY^T)\dot{Y}\right) - L(Y)\right] - \frac{1}{2}\mathrm{Tr}\left(\bar{A}^T(Y^TY - \mathbb{I})\right),
\end{equation}
where $a$ is a parameter that defines a \textit{one parameter family of Riemannian metrics} on the Stiefel manifold $\mathcal{M}$, i.e. $g_a:T_Y\mathcal{M}\times{}T_Y\mathcal{M} \to \mathbb{R}^+,\,(V_1, V_2)\mapsto\frac{1}{2}\mathrm{Tr}(V_1^T(I - aYY^T)V_2)$, and $L$ is the loss function to be minimized. In this description $Y$, which are elements of $\mathcal{M}$, are parameters of the neural network and $\bar{A}\in\mathbb{R}^{n\times{}n}$ is a Lagrange multiplier that enforces the constraint that the neural network weights have to be elements of the Stiefel manifold.

The authors obtain equations of motion through the variational principle and these are then discretized by a clever splitting scheme to obtain the final optimizer, which is used for training the neural network. Stochastic gradient descent (SGD) with momentum is in this case just the solution of the variational problem, which the authors call ``Momentum SGD''. In order to obtain a version of Adam, the authors apply a modification to this algorithm to include second moments.

In their approach the authors do not use the interpretation of the Stiefel manifold as a homogeneous space, i.e. the tangent space is not viewed as $T_Y\mathcal{M} = \mathfrak{g}\cdot{}Y$, but rather as a decomposition into a \textit{parallel} and a \textit{normal component}\footnote{We note that this splitting is different from the decomposition of $\mathfrak{g}$ into $\mathfrak{g}^{\mathrm{hor},Y}$ and $\mathfrak{g}^{\mathrm{ver},Y}$.}:

\begin{equation}
    \begin{aligned}
    T_Y\mathcal{M} & = (T_Y\mathcal{M})_\parallel\oplus(T_Y\mathcal{M})_\perp \\ & = \{YA:\text{$A\in\mathbb{R}^{n\times{}n}$ skew-sym.}\}\oplus\{\bar{B}\in\mathbb{R}^{N\times{}n}:\bar{B}^TY = \mathbb{O}\}
    \end{aligned}
    \label{eq:stiefel_tangent_decomposition_alternative}
\end{equation}

Note the difference between $\bar{B}$ in \Cref{eq:stiefel_tangent_decomposition_alternative} and $B$ in \Cref{eq:g_hor}. There is however a connection between $B$ and $\bar{B}$: for the case when $Y = E$ the $B$ part in \Cref{eq:g_hor} can be identified with $\bar{B}\in(T_E\mathcal{M})_\perp$ by writing:

\begin{equation}
    \bar{B} = \begin{pmatrix} \mathbb{O} \\ B \end{pmatrix}.
\end{equation}

In \cite[Algorithm 2]{kong2022momentum} the expression of $\mathrm{grad}_YL$ in this basis is given as:

\begin{equation}
    (A, \bar{B}) = (Y^T\mathrm{grad}_YL, \mathrm{grad}_YL - YY^T\mathrm{grad}_YL) = (Y^T\nabla{}L - (\nabla{}L)^TY, (\mathbb{I} - YY^T)\nabla{}L).
\end{equation}

The second moments of the Adam optimizer are then computed by taking $A\odot{}A$ and $\bar{B}\odot\bar{B}$. These operations however destroy the tangent space structure. Especially the second one is not easy to fix. This is acknowledged in \cite{kong2022momentum} and ``solved'' by applying a projection to the $\bar{B}$ part after it was updated: $\bar{B} \gets (\mathbb{I} - YY^T)\bar{B}$. We also mention that the application of our optimizer in \Cref{transformer} (i.e. training a transformer \cite{vaswani2017attention} with weights on the Stiefel manifold) was directly inspired by \cite{kong2022momentum}.

We note that for Lie groups (i.e. $n = N$, $\mathcal{M} = G$ and $\mathfrak{g}^\mathrm{hor} = \mathfrak{g}$) the optimizer we present in this work is equivalent to \cite[Algorithm 5]{kong2022momentum} and can be derived with the approach outlined in \cite{lezcano2019cheap}. In \cite{lezcano2019cheap} the authors design an optimization scheme for which a single update\footnote{As was already explained in \cite{lezcano2019cheap} the exponential map can be replaced with a \textit{retraction} \cite{absil2008optimization} (also confer \Cref{def:classical_retraction}).} is

\begin{equation}
    e^A \gets \exp(A - \eta\nabla(L\circ\exp)(A)).
\end{equation}

So in this approach the problem of updating the cache on $G$ is rephrased as a problem of updating the cache in the vector space $\mathfrak{g}$ by viewing $\exp:\mathfrak{g}\to{}G$ (or a retraction $\mathcal{R}:\mathfrak{g}\to{}G$) as a coordinate map. This is similar to how Runge-Kutta methods are extended to Runge-Kutta-Munthe-Kaas methods \cite{munthe1998runge}. Noting that this approach can generalize existing optimizers to Lie groups but not to homogeneous spaces, \cite{lezcano2019cheap} is concluded by saying: ``Additionally, it could be of interest to see how orthogonal constraints help with learning in deep feed forward networks. In order to make this last point formal, one would have to generalize the results presented here to homogeneous Riemannian manifolds, like the Stiefel manifold.'' In \Cref{general_framework} we show how we make this ``generalization''. 


\section{Generalizing Adam by using Global Sections}
\label{general_framework}

In this section we show how to generalize the Adam optimizer from \Cref{adam_section} to homogeneous spaces using the material discussed in \Cref{homogeneous_spaces}. Here we also discuss some of the computational aspects of the proposed algorithm to make clear how it can be implemented.

We show our high-level optimization algorithm (i.e. not specifically restricted to the \textit{Adam form}) for general neural networks in \Cref{alg:general_man}. This is an extension of \Cref{alg:adam_alg}. We further visualize the new optimizer framework (along with traditional vector space optimization) in \Cref{fig:new_optimizer}. In \Cref{fig:new_optimizer} we also indicate the use of AD to compute the Euclidean gradient $\nabla{}L$. This is then converted to a Riemannian gradient by using \texttt{rgrad} (see \Cref{op:rgrad}) in \Cref{alg:general_man}. In order to highlight the \textit{operations} that need to be implemented to extend a neural network optimizer to homogeneous spaces with the approach presented here, we explicitly define \Cref{op:rgrad,op:global_rep,op:retraction}. Besides these operations, the other steps are equivalent to standard optimizers in Euclidean space.
\begin{algorithm}
    \caption{High-level optimization algorithm for neural networks with manifolds}\label{alg:general_man}
    \begin{algorithmic}
    \Require time $t$, weights $Y^{(t)}$, \texttt{cache}, differential $\nabla{}L$, $\mathcal{O}$-parameters $\Xi$, section $\Lambda^{(t)}$;
    \State $t \gets t + 1$
    \State $\Delta^{(t)} \gets \mathtt{rgrad}(\nabla{}L)$ \Comment{Riemannian gradient (\Cref{op:rgrad})}
    \State $\mathcal{B}^{(t)} \gets \mathtt{global\_rep}(Y^{(t)}, \Delta^{(t)})$ \Comment{Global representation (\Cref{op:global_rep})}
    \State $\mathtt{cache} \gets \mathtt{update}(\mathtt{cache}, \mathcal{B}^{(t)}, t, \Xi)$ \Comment{Update the cache (\Cref{dftn:method})}
    \State $W^{(t)} \gets \mathtt{velocity}(\mathtt{cache}, \Xi)$ \Comment{Final velocity (\Cref{dftn:method})}
    \State $\Lambda^{(t+1)} \gets \mathtt{update\_section}(\Lambda^{(t)}, W^{(t)})$ \Comment{Section update (1st part of \Cref{op:retraction})}
    \State $Y^{(t+1)} \gets \mathtt{apply\_section}(\Lambda^{(t+1)}, E)$ \Comment{Multiply with $E$ (2nd part of \Cref{op:retraction})}
    \end{algorithmic}
\end{algorithm}

We stress that the distinguishing feature of our optimizer, and the one that sets it apart from e.g. \cite{li2020efficient,kong2022momentum}, is that we always work with spaces of the same dimension: $\mathrm{dim}St(n, N) = \mathrm{dim}\mathfrak{g}^{\mathrm{hor},Y} = \mathrm{dim}\mathfrak{g}^\mathrm{hor} = n(N - n) + n(n-1)/2$ and therefore do not need to perform a projection. This is different in e.g. \cite{kong2022momentum} (see \Cref{existing_approaches}) as the authors there work on spaces $\mathbb{R}^{N\times{}n}\times\{A\in\mathbb{R}^{n\times{}n}:\text{$A$ is skew-sym.}\}$, which has dimension $Nn + n(n-1)/2$.

For the case for which the weights lie on a vector space $\mathcal{V}$ (a special case of a homogeneous space), the functions \texttt{rgrad}, \texttt{global\_rep}, \texttt{update\_section} and \texttt{apply\_section} (we further group \texttt{update\_section} and \texttt{apply\_section} together under $\overline{\mathrm{retraction}}$\footnote{As is shown in \Cref{fig:new_optimizer} $\overline{\mathrm{retraction}}$ is the composition of \texttt{update\_section} and \texttt{apply\_section}. We call $\overline{\mathrm{retraction}}$ an \textit{extended retraction} (see \Cref{op:retraction}).}) are greatly simplified and we recover \Cref{alg:adam_alg}. Specifically:

\begin{itemize}
    \item \texttt{rgrad} and \texttt{global\_rep} are identity mappings.
    \item $\overline{\mathrm{retraction}}$ is \textit{standard addition}. Note that in \Cref{fig:new_optimizer} and \Cref{alg:general_man} we split $\overline{\mathrm{retraction}}$ up into two mappings: \texttt{update\_section} and \texttt{apply\_section}. When we deal with a vector space $\mathcal{V}$, \texttt{update\_section} is the identity and \texttt{apply\_section} is addition.
    \item the \textit{distinct element} $E\in\mathcal{M}$ is the zero element $\mathbb{O}\in\mathcal{V}$. The special role of this distinct element is explained in \Cref{homogeneous_spaces}.
\end{itemize}

\begin{figure}

    {

    \centering

    \begin{tikzpicture}
    \node[rectangle, draw] (R1)    at (-.25,0) {$\mathbb{R}^N$};
    \node[rectangle, draw] (R2)  at (2, 3) {$\mathbb{R}^N$};
    \node[rectangle, draw] (cache) at (2, 0) {$\mathtt{cache}$};
    \node[rectangle, draw] (R3) at (5.5, 1.8) {$\mathbb{R}^N$};
    \node[rectangle, draw] (R4)	at (5.5, 0) {$\mathbb{R}^N$};
    \coordinate[right of=R3, xshift=.6cm] (addition);
    \node[fit=(R2)(cache)(R3)(R4)(addition),draw, ultra thick, rounded corners, label=below:$\mathtt{optimization\_step(o, ::}\mathbb{R}^N\mathtt{,::}\mathbb{R}^N\mathtt{)}$] (optimization_step) {};

    \draw[->] (R2) -- (cache) node[pos=.5, sloped, below] {\small\texttt{update!}};
    \draw[->] (cache) -- (R3) node[pos=.5, sloped, above] {\small\texttt{velocity}};
    \draw[->] (R3) -- (R4) node[pos=.5, right, xshift=-.1cm] {\small Addition};
    \draw[->, mgreen] (R1) -- (R2) node[pos=.5, left] {\small\color{mgreen}AD};
\end{tikzpicture}\hspace{.2cm}\vline\hspace{.2cm}\begin{tikzpicture}
    \node[rectangle, draw] (TYM)   at (0, 0) {$T_Y\mathcal{M}$};
    \node[rectangle, draw] (ghorY) at (0, 2.1) {$\mathfrak{g}^{\mathrm{hor},Y}$};
    \node[rectangle, draw] (ghor)  at (2, 2.1) {$\mathfrak{g}^{\mathrm{hor}}$};
    \node[rectangle, draw] (cache) at (2, 0) {$\mathtt{cache}$};
    \node[rectangle, draw] (ghor2) at (4, 1) {$\mathfrak{g}^\mathrm{hor}$};
    \node[rectangle, draw] (M)	   at (4, 0) {$G$};
    \node[rectangle, draw] (Euc)   at (-2.4,0) {$\mathbb{R}^{N\times{}n}$};
    \node[rectangle, draw] (M2)    at (-3.8,-1) {$\mathcal{M}$};
    \node[rectangle, draw] (M3)    at (4, -1) {$\mathcal{M}$};
    \coordinate[right of=M, xshift=1.95cm] (apply_section);

    \draw[->] (TYM) -- (ghorY) node[pos=.5, left] {$\Omega$};
    \draw[->, mred] (ghorY) -- (ghor);
    \draw[->] (ghor) -- (cache) node[pos=.5, sloped, below] {\small\texttt{update!}};
    \draw[->] (cache) -- (ghor2) node[pos=.5, sloped, above] {\small\texttt{velocity}};
    \draw[->, mred] (ghor2) -- (M) node[pos=.5, right] {\small $\mathtt{update\_section}$};
    \draw[->, mgreen] (M2) -- (Euc) node[pos=.5, left] {\small\color{mgreen}AD};
    \draw[->] (Euc) -- (TYM) node[pos=.5, above, yshift=-.15cm] {\small$\mathtt{rgrad}$};
    \draw[->, mred] (M) -- (M3) node[pos=.5, right, xshift=-.1cm] {\small $\mathtt{apply\_section}$};

    \coordinate[above of=TYM, yshift=-.27cm] (above_of_TYM);
    \coordinate[left of=ghor, xshift=.23cm] (left_of_ghor);
    \coordinate[below of=ghor2, yshift=.63cm] (below_of_ghor2);
    \coordinate[above of=M3, yshift=-.68cm] (above_of_M);
    \node[fit=(above_of_TYM)(ghorY)(left_of_ghor), label={[rotate=90, yshift=.2cm, xshift=1.05cm]left:{\color{morange!70}{\small$\mathtt{global\_rep}$}}}, draw, morange!70, rounded corners] (global_rep) {};
    \node[fit=(below_of_ghor2)(above_of_M)(M)(apply_section), label={[xshift=.3cm]above:{\color{morange!70}{\small{}$\overline{\mathrm{retraction}}$}}}, draw, morange!70, rounded corners] (retraction) {};
    \node[fit=(Euc)(ghorY)(ghor)(ghor2)(M3)(apply_section)(global_rep)(retraction),draw, ultra thick, rounded corners, label=below:$\mathtt{optimization\_step!(o::Optimizer,::}G\mathcal{, ::}\mathcal{M}\mathtt{,::}\mathbb{R}^{N\times{}n}\mathtt{)}$] (optimization_step) {};
\end{tikzpicture}

    }

    \caption{Comparison between regular Adam (shown in \Cref{alg:adam_alg}) and Adam adapted to manifolds (shown in \Cref{alg:general_man}). A red color indicates that the operation requires the \textit{global section} $\Lambda^{(t)}$. An orange color indicates that the operation is the result of grouping two individual operations together. With the framework presented in this work we can recover the optimizer shown on the left from the one on the right if the manifold $\mathcal{M}$ is a vector space.}
    \label{fig:new_optimizer}
\end{figure}

The other mappings, i.e. $\mathtt{update!}$ and $\mathtt{velocity}$, as well as the structure of the \texttt{cache} and the parameters of the optimizer (called $\mathcal{O}$-parameters in \Cref{alg:general_man}), are however equivalent in both cases (in \Cref{transformer} we explicitly list the $\mathcal{O}$-parameters we use). Note that we already defined an optimizer method in \Cref{dftn:method} as the combination of an \texttt{update!} method, a \texttt{velocity} method and a \texttt{cache}.

\Cref{alg:general_man} comprises generalizations of all common first-order optimization algorithms such as (stochastic) gradient descent (with and without momentum), Adam, RMSProp and AdaGrad.  The mappings \texttt{rgrad}, \texttt{global\_rep}, \texttt{update\_section} and \texttt{apply\_section}, which are needed in addition to the usual steps performed by an optimizer for vector spaces, are now discussed in detail.

\begin{operation}[\texttt{rgrad}]\label{op:rgrad}
This function computes the Riemannian gradient of a loss function $L$ on a Riemannian matrix manifold $\mathcal{M}\subset\mathbb{R}^{N\times{}n}$ based on its Euclidean gradient $\nabla{}L\in\mathbb{R}^{N\times{}n}$.
\end{operation}

For matrix manifolds with Riemannian metric $g$, \texttt{rgrad} can always be found since the Euclidean gradient corresponds to an element of the cotangent space $T_Y^*\mathcal{M}$ and every element of $T_Y^*\mathcal{M}$ can be converted to an element of $T_Y\mathcal{M}$ with a Riemannian metric (see \cite[Chapter 5]{bishop1980tensor}):
\begin{equation}
    \langle{}dL,V\rangle \equiv  \mathrm{Tr}((\nabla_YL)^TV) = g_Y(\mathtt{rgrad}(Y,\nabla_YL), V),\quad\forall{}V\in{}T_Y\mathcal{M}.
    \label{eq:rgrad}
\end{equation}

For the Stiefel manifold (see \Cref{homogeneous_spaces}) the canonical metric and the associated gradient are \cite[Equation (2.39)]{edelman1998geometry}:

\begin{equation}
    g_Y(V_1, V_2) = \mathrm{Tr}\left(V_1^T\left(\mathbb{I} - \frac{1}{2}YY^T\right)V_2\right),
    \label{eq:stiefel_manifold_metric}
\end{equation}

and \cite[Equation (2.53)]{edelman1998geometry}

\begin{equation}
    \mathrm{grad}_YL = \mathtt{rgrad}(Y, \nabla{}L) = \nabla{}L - Y\nabla{}L^TY.
    \label{eq:stiefel_manifold_rgrad}
\end{equation}

For homogeneous spaces there is a natural way of obtaining a metric and this metric is hence called the \textit{canonical metric}. How to obtain \Cref{eq:stiefel_manifold_metric} with the help of \Cref{eq:omega} was mentioned in \Cref{homogeneous_spaces}. After we have applied \texttt{rgrad}, we get an element in $T_Y\mathcal{M}$. We cannot use this for updating the \texttt{cache} however, since the parametrization of this tangent vector depends on the specific tangent space $T_Y\mathcal{M}$ (also see the discussion in \Cref{introduction}). In order to get around this issue we use \textit{global tangent space representations} (confer \Cref{def:global_tangent_space}). We now describe the operation \texttt{global\_rep} in \Cref{alg:general_man}:

\begin{operation}[\texttt{global\_rep}]\label{op:global_rep}
    A mapping from the tangent space $T_Y\mathcal{M}$, the output of \texttt{rgrad}, to the \textit{global tangent space} $\mathfrak{g}^{\mathrm{hor}} \simeq T_Y\mathcal{M}$.
\end{operation}

So \texttt{global\_rep} realizes one direction of the isomorphism $T_Y\mathcal{M} \overset{\sim}{\rightarrow}\mathfrak{g}^\mathrm{hor}$, i.e. every element of the tangent space $T_Y\mathcal{M}$, for every $Y\in\mathcal{M}$, can be \textit{lifted} to $\mathfrak{g}^\mathrm{hor}$. For computational purposes \texttt{global\_rep} performs two steps simultaneously as indicated in \Cref{fig:new_optimizer}: the first map is \Cref{eq:omega} and the second one is $\mathfrak{g}^\mathrm{hor,Y}\to\mathfrak{g}^\mathrm{hor}$. This second map uses global sections (see \Cref{homogeneous_spaces}).

\begin{remark}
    If $\mathcal{M}$ is a vector space $\mathcal{V}$, then $\mathcal{M} \equiv  T_Y^*\mathcal{M} \equiv T_Y\mathcal{M} \equiv \mathfrak{g}^\mathrm{hor} = \mathcal{V}$ and no \textit{global tangent space representation} is necessary.
\end{remark}

\begin{remark}
    In the implementation we do not store an $N\times{}N$ matrix for $\mathcal{B}\in\mathfrak{g}^\mathrm{hor}$, but simply an $n\times{}n$ skew-symmetric matrix $A$ (parametrized as an $n(n-1)$-valued vector internally) and an arbitrary $(N-n)\times{}n$ matrix $B$. The operation \texttt{global\_rep} then simply computes $A = Y^T\Delta$ and $B = Y_\perp^T\Delta$, where $Y_\perp$ is the orthogonal complement to $Y$, i.e. $\Lambda = \lambda(Y) = [Y, Y_\perp]$.
\end{remark}

The operation $\overline{\mathrm{retraction}}$ (that is split into the operations \texttt{update\_section} and \texttt{apply\_section} in \Cref{alg:general_man} and \Cref{fig:new_optimizer}) is a slight modification of a \textit{classical retraction} (see \Cref{def:classical_retraction}):

\begin{operation}[$\overline{\mathrm{retraction}}$]\label{op:retraction}
    We call a map from the global tangent space to the manifold $\overline{\mathrm{retraction}}: \mathfrak{g}^\mathrm{hor} \to \mathcal{M}$ an \textbf{extended retraction} if $\overline{\mathrm{retraction}}\circ\Omega:T_E\mathcal{M}\to\mathcal{M}$ is a classical retraction according to \Cref{def:classical_retraction} with $\Omega$ being the unique identification $T_E\mathcal{M}\to\mathfrak{g}^\mathrm{hor}$.
\end{operation}

\begin{remark}
Since the Stiefel manifold admits closed-form solutions of its geodesic (see \Cref{eq:geodesic}) we can use these directly besides other approximations. Consequently we refer to the analytic solution of the geodesic spray as the \textit{geodesic retraction}. Besides the exponential retraction, we also treat the Cayley retraction \cite{absil2008optimization, li2020efficient} (see \Cref{cayley}) in more detail in this work.
\end{remark}

\begin{remark}
    If the considered manifold is a vector space $\mathcal{V}$ with the canonical metric, the geodesic is simply a straight line: $\gamma(t) = Y + tW\in\mathcal{V}$ for $Y, W \in\mathcal{V}$; i.e. the retraction is simple addition (also confer \Cref{fig:new_optimizer}).
\end{remark}

\begin{figure}
    \centering
    \begin{tikzpicture}[
    node1/.style={ellipse, draw=black!60, fill=mgreen!25, very thick, minimum height=17mm},
    node2/.style={ellipse, draw=black!60, fill=mblue!25, very thick, minimum height=17mm},
    arrow/.style={-Latex, very thick, rounded corners}
    ]

\node[node1, align = center] (extended) {\textbf{Extended Retraction}\\ $\overline{\mathrm{retraction}}:\mathfrak{g}^\mathrm{hor}\to\mathcal{M}$ \\ comprises \texttt{update\_section} and \texttt{apply\_section}};
\node[node2, right of=extended, xshift = 10cm, align = center] (classical) {\textbf{Classical Retraction} \\ $T_E\mathcal{M}\to\mathcal{M}$};

\draw[arrow] (extended) to node[sloped, midway, above] {$\overline{\mathrm{retraction}}\circ\Omega$} (classical);

\end{tikzpicture}
    \caption{The \textit{extended retraction} (see \Cref{op:retraction}) $\overline{\mathrm{retraction}}:\mathfrak{g}\to\mathcal{M}$ we use in this work is different from \textit{classical retractions} (see \Cref{def:classical_retraction}) but fulfills a similar role. An \textit{extended retraction} is defined as a map from which a \textit{classical retraction} can be recovered by composing it with another operation.}
    \label{fig:retraction_rep}
\end{figure}

The reason for why the actual computation of the geodesic is split up into two parts, called \texttt{update\_section} and \texttt{apply\_section} in \Cref{fig:new_optimizer} can be seen in \Cref{eq:geodesic}, whose geodesic direction is the final velocity $W^{(t)}\in\mathfrak{g}^{\mathrm{hor}}$ of \Cref{alg:general_man}. We first compute $\Lambda^{(t)}\exp(W^{(t)})$ (called \texttt{update\_section}) and then perform a right multiplication by $E$ (called \texttt{apply\_section}). If we use a retraction instead of the exponential map we take
\begin{equation}
    \mathtt{update\_section}(\Lambda^{(t)}, W^{(t)}) = \Lambda^{(t)}\mathrm{retraction}(W^{(t)}).
\end{equation}

In \Cref{transformer} we compare two retractions: the geodesic retraction and the Cayley retraction. In \Cref{computing_the_exponential,cayley} we show how to utilize the sparse structure of $\mathfrak{g}^\mathrm{hor}$ (see \Cref{eq:g_hor}) to make the computation of the retraction especially efficient. Similar sparse representations of geodesics and the Cayley transform can be found in \cite{bendokat2021real, gao2024optimization, fraikin2007optimization, celledoni2000approximating}.

We also note that for a geometrical sound optimization scheme the cache should be \textit{parallel transported along the optimization trajectory} \cite{absil2008optimization, ring2012optimization}. In \Cref{parallel_transport} we give a short discussion of parallel transport for the Stiefel manifold.

Now recalling the contents of \Cref{adam_section}, we can see how to obtain the generalized Adam from \Cref{alg:general_man}: simply have to use \Cref{eq:adam_cache,eq:adam_velocity}. The $\mathcal{O}$-parameters (see \Cref{alg:general_man}) for Adam are $\Xi = (\eta, \beta_1, \beta_2, \delta)$. The Adam \texttt{cache} and the Adam operations in \Cref{eq:adam_cache,eq:adam_velocity} are identical in the vector space and in the manifold case.

For the case $N=n$, the versions of Adam presented in \cite[Algorithm 5]{kong2022momentum} and \cite{lezcano2019cheap} are very similar to our approach. In this case the homogeneous space is just the Lie group and the \textit{global tangent space} is the associated Lie algebra $\mathfrak{g}$. In this case no lift has to be performed. But if $N\neq{}n$, then \cite{kong2022momentum} need a projection, which our algorithm does not need and the approach shown in \cite{lezcano2019cheap} is not able to deal with the case $N\neq{}n$ at all; it can only be applied to Lie groups and not to homogeneous spaces.

In \cite{kong2022momentum} the authors write regarding $\mathcal{M} = SO(N)$: ``Note [that Adam] can be derived from either the special case of the Stiefel optimizer, or the structure of Lie algebra [sic] $\mathfrak{so}(n)$, i.e., left-trivialization [...] such that the momentum [...] can be recognized as an element in $T_\mathbb{I}SO(N)$, the tangent space at the identity. In the latter interpretation, the momentum will always stay in the same tangent space $T_\mathbb{I}SO(N)$ which passes on a message that trivialization almost reproduce [sic] the convenience in Euclidean space for $SO(N)$. Hence unlike the general Adam-Stiefel optimizer, no extra technique, for example projection, is needed for Adam-$SO(N)$.'' By ``general Adam-Stiefel optimizer'' the authors refer to their \textit{generalization of Adam} discussed in \Cref{existing_approaches}. In this work we now made the additional step of ``reproducing the convenience in Euclidean space'' not just for the Lie groups like $SO(N)$ but also for the homogeneous spaces like $St(n, N)$.

In \Cref{transformer} we compare Adam to two other optimizers: (i) the \textit{gradient optimizer} and (ii) the \textit{momentum optimizer}. Both can be derived from the general framework presented in \Cref{alg:adam_alg,alg:general_man}. For these two optimizers:

\begin{enumerate}
    \item[(i)] \texttt{cache} is empty; the final velocity $W^{(t)}$ in \Cref{alg:general_man} is set to $-\eta\mathcal{B}^{(t)}$ and $\Xi = (\eta)$.
    \item[(ii)] \texttt{cache} stores velocity information, i.e. consists of first moments $\mathcal{B}^\mathtt{cache}$. This is updated with
    \begin{equation}
        \mathcal{B}^\mathtt{cache} \gets \alpha\mathcal{B}^\mathtt{cache} + \mathcal{B}^{(t)}.
    \end{equation}
    The velocity is then computed with $W^{(t)}\gets \mathtt{velocity}(\mathtt{cache}, \Xi) := -\eta\mathcal{B}^\mathtt{cache}$, where $\Xi = (\eta, \alpha)$.
\end{enumerate}

\subsection{Decoupled Weight Decay and the Absence of a Manifold AdamW}
\label{weight_decay}

Nowadays most researchers do not use the standard Adam optimizer but modified versions, most prominently AdamW \cite{loshchilov2017decoupled}, i.e. Adam with \textit{decoupled weight decay}. Here we briefly discuss what this variant becomes in the framework in \Cref{alg:general_man}, and the answer is unusually clean: on the compact manifolds relevant here it becomes Adam again.

AdamW works by introducing ordinary $L^2$ regularization, i.e. adding the term $\Sigma(Y) = \frac{1}{2}||Y||_F^2$ in \Cref{eq:regularized_loss}, where $||\cdot||_F$ denotes the Frobenius norm. The associated term $\mu{}Y$ is added to the final velocity:

\begin{equation}
    W^{(t)} \gets -\eta\mathcal{B}_1^\mathtt{cache}/\sqrt{\mathcal{B}_2^\mathtt{cache} + \delta} - \eta\mu{}Y^{(t)}.
    \label{eq:decoupled_weight_decay}
\end{equation}

Note that $\mu{}Y^{(t)}$ never enters $\mathcal{B}_1^\mathtt{cache}$ or $\mathcal{B}_2^\mathtt{cache}$ and that this is a decay of the \textit{weights}.

The idea behind \textit{decoupled weight decay} cannot be carried over to \Cref{alg:general_man}; the final velocity there is an element of the global tangent space $\mathfrak{g}^\mathrm{hor}$, whereas $\mu{}Y^{(t)}$ has no geometric meaning; it is just an element of $\mathbb{R}^{N\times{}n}$. One can work around this by taking the Riemannian gradient of the $L^2$ regularizer, but this is zero as, by definition, $Y$ is an element of a compact space and does not decay\footnote{The same holds for other unitarily invariant norms, such as the spectral or the nuclear norm, which are likewise constant on $St(n, N)$.}. This holds for the Stiefel manifold as well as the Grassmann manifold \cite{bendokat2020grassmann}.

We further note that there is a different recent approach \cite{gu2026mano} that relaxes restrictions on the weights $Y^{(t)}$ and allows them to drift away from the manifold, thus incorporating weight decay; but this approach also introduces an additional projection step that is not necessary in the approach presented here.

For a network whose weights are of mixed type, i.e. Stiefel weights alongside unconstrained weights and biases, as in the transformer of \Cref{transformer}, the practical consequence is that decoupled weight decay applies to the unconstrained weights exactly as in \cite{loshchilov2017decoupled} and to the manifold weights not at all\footnote{This is what the implementation of AdamW in the \texttt{Julia} package \texttt{GeometricOptimizers.jl} \cite{brantner2026geometricoptimizers} does: on a bare manifold the method coincides with Adam.}.



%
\newcommand{\mnistSeeds}{10}
\newcommand{\mnistEpochs}{500}
\newcommand{\mnistTotalRuns}{51}
\newcommand{\mnistExcludedRuns}{1}
\newcommand{\mnistBackend}{cuda}
\newcommand{\mnistGeometricAccepted}{10}
\newcommand{\mnistGeometricAccuracy}{$86.02 \pm 0.57$}
\newcommand{\mnistBaselineAccuracy}{$76.20 \pm 0.75$}
\newcommand{\mnistAccuracyGain}{9.8}
\newcommand{\mnistGeometricWorstAccuracy}{84.90}
\newcommand{\mnistBaselineBestAccuracy}{77.25}
\newcommand{\mnistGradientMs}{118}
\newcommand{\mnistDirectionMs}{4.8}
\newcommand{\mnistRetractionMs}{20.8}
\newcommand{\mnistGeometricShare}{18}
\newcommand{\mnistStepOverhead}{11}
\newcommand{\mnistEpochOverhead}{7}
\newcommand{\mnistInstrumentedShare}{54}
\newcommand{\mnistUnattributedShare}{46}
\newcommand{\mnistPeakMemory}{21.8}
\newcommand{\mnistGeometricStateMiB}{4.34}
\newcommand{\mnistAblationStateMiB}{1.77}
\newcommand{\mnistBaselineStateMiB}{3.94}
\newcommand{\mnistStateFraction}{0.019}
\newcommand{\mnistHorizontalEntries}{315}
\newcommand{\mnistAmbientEntries}{343}
\newcommand{\mnistAlgebraEntries}{1176}
\newcommand{\mnistGlobalSectionEntries}{2401}
\newcommand{\mnistStiefelWeights}{336}
\newcommand{\mnistTargetLoss}{0.344}
\newcommand{\mnistTargetEpochs}{155}
\newcommand{\mnistTargetMinutes}{20}
\newcommand{\mnistTargetReached}{10}
\newcommand{\mnistBaselineMinutes}{47}
\newcommand{\mnistDriftMax}{$9.0{\times}10^{-3}$}
\newcommand{\mnistDriftMin}{$8.5{\times}10^{-3}$}
\newcommand{\mnistBaselineDrift}{$8.4{\times}10^{-4}$}
\newcommand{\mnistDriftRatio}{19}
\newcommand{\mnistDriftTolerance}{$10^{-2}$}
\newcommand{\mnistDriftHeadroom}{10}
\newcommand{\mnistAblationLoss}{1.337}

%
\newcommand{\fashionSeeds}{10}
\newcommand{\fashionEpochs}{500}
\newcommand{\fashionTotalRuns}{53}
\newcommand{\fashionExcludedRuns}{3}
\newcommand{\fashionBackend}{cuda}
\newcommand{\fashionGeometricAccepted}{10}
\newcommand{\fashionGeometricAccuracy}{$78.66 \pm 0.51$}
\newcommand{\fashionBaselineAccuracy}{$75.35 \pm 0.62$}
\newcommand{\fashionAccuracyGain}{3.3}
\newcommand{\fashionGeometricWorstAccuracy}{77.55}
\newcommand{\fashionBaselineBestAccuracy}{76.30}
\newcommand{\fashionGradientMs}{117}
\newcommand{\fashionDirectionMs}{4.9}
\newcommand{\fashionRetractionMs}{20.8}
\newcommand{\fashionGeometricShare}{18}
\newcommand{\fashionStepOverhead}{10}
\newcommand{\fashionEpochOverhead}{6}
\newcommand{\fashionInstrumentedShare}{54}
\newcommand{\fashionUnattributedShare}{46}
\newcommand{\fashionPeakMemory}{22.2}
\newcommand{\fashionGeometricStateMiB}{4.34}
\newcommand{\fashionAblationStateMiB}{1.77}
\newcommand{\fashionBaselineStateMiB}{3.94}
\newcommand{\fashionStateFraction}{0.019}
\newcommand{\fashionHorizontalEntries}{315}
\newcommand{\fashionAmbientEntries}{343}
\newcommand{\fashionAlgebraEntries}{1176}
\newcommand{\fashionGlobalSectionEntries}{2401}
\newcommand{\fashionStiefelWeights}{336}
\newcommand{\fashionTargetLoss}{0.354}
\newcommand{\fashionTargetEpochs}{290}
\newcommand{\fashionTargetMinutes}{37}
\newcommand{\fashionTargetReached}{10}
\newcommand{\fashionBaselineMinutes}{45}
\newcommand{\fashionDriftMax}{$8.9{\times}10^{-3}$}
\newcommand{\fashionDriftMin}{$8.6{\times}10^{-3}$}
\newcommand{\fashionBaselineDrift}{$3.9{\times}10^{-4}$}
\newcommand{\fashionDriftRatio}{29}
\newcommand{\fashionDriftTolerance}{$10^{-2}$}
\newcommand{\fashionDriftHeadroom}{11}
\newcommand{\fashionAblationLoss}{1.342}

%
\newcommand{\pendulumSeeds}{10}
\newcommand{\pendulumEpochs}{1000}
\newcommand{\pendulumTotalRuns}{40}
\newcommand{\pendulumExcludedRuns}{0}
\newcommand{\pendulumBackend}{cuda}
\newcommand{\pendulumGeometricFinal}{$0.292 \pm 0.055$}
\newcommand{\pendulumBaselineFinal}{$0.394 \pm 0.072$}
\newcommand{\pendulumGeometricBest}{$0.281 \pm 0.059$}
\newcommand{\pendulumBaselineBest}{$0.289 \pm 0.050$}
\newcommand{\pendulumFinalReduction}{26.0}
\newcommand{\pendulumGeometricWins}{9}
\newcommand{\pendulumGeometricMinutes}{$17.84 \pm 0.05$}
\newcommand{\pendulumBaselineMinutes}{$17.91 \pm 0.05$}

\section{Numerical Example: the Transformer}
\label{transformer}

The transformer architecture \cite{vaswani2017attention}, originally conceived for natural language processing tasks, and the vision transformer \cite{dosovitskiy2020image} (used for processing image data) have largely driven advances in these fields in recent years.
Part of the transformer architecture's allure is that it is a simple and elegant construction that makes the interpretation of the neural network possible, generalizes well to diverse data sets and, perhaps most importantly, its optimization is easily parallelizable which implies good performance on GPUs.

Despite all of this, caution has to be taken when training a transformer network. The original vision transformer paper \cite{dosovitskiy2020image}, for example, relies on techniques such as layer normalization \cite{ba2016layer} and extensive pre-training.  Here we demonstrate that this is not necessary when optimizing on the Stiefel manifold. We also note that constrained optimization on the Stiefel manifold to train a vision transformer was already done in \cite{kong2022momentum}. Also, it has been shown in \cite{zhang2021orthogonality} that weakly enforcing orthonormality constraints, i.e. adding terms that penalize violations of orthonormality constraints via a regularizer as shown in \Cref{eq:regularized_loss}, are of advantage when dealing with transformers\footnote{Orthogonality constraints thus conform to \textit{weakly enforcing the Stiefel property}. Note that weakly enforcing constraints requires an additional hyperparameter and this can make training difficult.}. 

\begin{figure}

    {

    \centering

    \includegraphics[width=.8\textwidth]{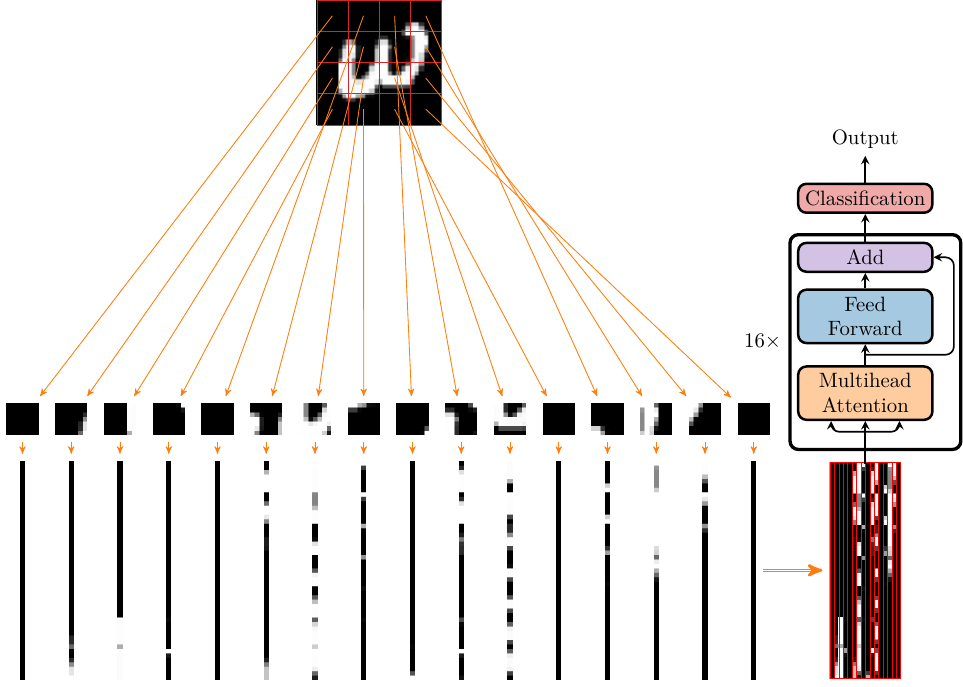}

    }

    \caption{Preprocessing of the images is done in two steps. First the image is split into 16 patches of size $7\times7$. In the second step the images are flattened and then put together again, yielding a $49\times16$ matrix. The output is then fed into the transformer.}
    \label{fig:preprocessing}
\end{figure}

We train the neural network on the MNIST data set \cite{deng2012mnist} and the Fashion-MNIST data set \cite{xiao2017fashion}: for both the training data each consist of 60000 labelled $(28\times28)$ images and the test data  consist of 10000 labelled $(28\times28)$ images. In order to apply the vision transformer, the data are split into 16 $(7\times7)$ image patches (similar to what was done in \cite{dosovitskiy2020image}), and these collections of 49-dimensional vectors are directly fed into the transformer. The preprocessing steps and the resulting input to the transformer (a $49\times16$ matrix) are shown in \Cref{fig:preprocessing}. With the notation used in \Cref{homogeneous_spaces} we have $N = 49$ and $n = 7$. The optimizers have been implemented as part of the \texttt{Julia} package \texttt{GeometricOptimizers.jl} \cite{brantner2026geometricoptimizers} and the training scripts as part of \texttt{GMLDatasets.jl}\footnote{ For the GPU implementation \texttt{CUDA.jl} \cite{besard2018juliagpu} is used.}, and all the training is done on an NVIDIA Geforce RTX 4090.

The right part of \Cref{fig:preprocessing} shows the transformer architecture. Its most important part are ``multihead attention modules'' (introduced by \cite{vaswani2017attention}). Taking as input a matrix $I\in\mathbb{R}^{49\times16}$ (the result of preprocessing an image), a single multihead attention layer performs three computations:

\begin{enumerate}
    \item[(i)] Computing three projections (via the matrices $W^Q_i$, $W^K_i$ and $W^V_i$) for each index $i$. The index $i=1,\ldots,7$ indicates the number of the ``attention head'' (here we have seven heads). The resulting matrices are called the ``queries'', ``keys'' and ``values'' in \cite{vaswani2017attention}.
    \item[(ii)] Weighting the projected \textit{values} $V_i := W^V_iI$ by the \textit{queries} $Q_i := W^Q_iI$ and the \textit{keys} $K_i := W^K_iI$ via an attention mechanism:

    \begin{equation}
        \hat{V}_i := \mathtt{Attention}(Q_i, K_i, V_i) = V_i\mathrm{softmax}(Q_i^TK_i),
        \label{eq:attention}
    \end{equation}
    where the softmax (see \Cref{softmax}) is computed column-wise.

    \item[(iii)] Rearranging the seven smaller matrices $\in\mathbb{R}^{7\times16}$, i.e. the seven outputs of the previous step, into a bigger matrix $\in\mathbb{R}^{49\times16}$ by concatenating them, i.e. the output of the multihead attention layer is $\begin{bmatrix} \hat{V}_1^T & \cdots & \hat{V}_7^T  \end{bmatrix}^T$.
\end{enumerate}

So in step (i) the 49-dimensional vectors of the input matrix (see \Cref{fig:preprocessing}) are projected to seven-dimensional vectors with three different projection matrices to obtain \textit{values}, \textit{queries} and \textit{keys}. In step (ii) correlations between the queries and keys are computed via scalar products and these are used to calculate a reweighting of the values.

In our experiments the matrices $W^Q_i$, $W^K_i$ and $W^V_i$ in the multihead-attention layers are constrained to lie on the Stiefel manifold (except for the \textit{standard Adam ablation}, whose weights are unconstrained). See \Cref{mha_correlation} for a further theoretical motivation for this. After applying the multihead attention layer as a preprocessing step, the output is fed into a \textit{single-layer feedforward neural network with an add connection}, i.e.

\begin{equation}
    \mathtt{feedforward}: x\mapsto{}x + \sigma(Ax +b).
\end{equation}

For the feedforward layer all weights are unconstrained, i.e. $A\in\mathbb{R}^{49\times49}$ and $b\in\mathbb{R}^{49}$. In our neural network we use 16 such transformer layers (as indicated in \Cref{fig:preprocessing}). The output of the last transformer is then finally fed into a classification layer that does:

\begin{equation}
    \mathtt{classification}: \mathbb{R}^{49\times16}\to\mathbb{R}^{10}, X \mapsto \mathrm{softmax}(WX\mathtt{[1:49,end]}).
\end{equation}

This layer takes the last column of the output matrix, multiplies it with a learnable matrix $W\in\mathbb{R}^{10\times{}49}$ and composes it with a softmax.

For the actual \textit{training of the transformer} we compare five different cases: (i) the transformer with all projection matrices on $\mathbb{R}^{49\times7}$ and trained with Adam (i.e. the \textit{standard Adam ablation}); putting all projection matrices on the Stiefel manifold and optimizing with (ii) gradient descent (iii) gradient descent with momentum (iv) Adam and (v) the Riemannian Adam of \cite{li2020efficient}, whose second moment is a single scalar per weight. We also call (ii) the \textit{gradient optimizer} and (iii) the \textit{momentum optimizer} (see \Cref{general_framework}), and we refer to (v) as the \textit{scalar-moment baseline}. Case (i) is the standard Adam ablation and case (v) a published baseline to be compared to the optimizer developed here.

The $\mathcal{O}$-parameters $\Xi$ (see \Cref{alg:general_man}) are set to the following values:

\begin{table}[tbph]
    \centering%
    \begin{tabular}{|lllll|}
        \toprule%
        Gradient optimizer:& $\eta = 0.001$. &                  &                   & \\
        Momentum optimizer:& $\eta = 0.001$,  & $\alpha = 0.5$.  &                   & \\
        Adam optimizer:    & $\eta = 0.001$, & $\beta_1 = 0.9$, & $\beta_2 = 0.99$, & $\delta=10^{-8}$.\\
        Scalar-moment Adam:& $\eta = 0.001$, & $\beta_1 = 0.9$, & $\beta_2 = 0.99$, & $\delta=10^{-8}$.\\
        \botrule%
    \end{tabular}
\end{table}

These particular values for the Adam optimizer are typically used as a default and usually do not need much tuning \cite[chapter 8]{goodfellow2016deep}. The batch size is set to 2048\footnote{This batch size is chosen so that the GPU is used as much as possible (see \Cref{tab:mnist_cost}).} and the network is trained for 500 epochs. The data as well as all network parameters are in single precision. For all experiments the regular neural network weights (i.e. the ones in the feedforward neural network and in the classification layer) are initialized with \textit{Glorot uniform} \cite{pmlr-v9-glorot10a}. The weights that belong to the Stiefel manifold are initialized according to \Cref{alg:Stiefel_initialization}. The probability distributions $\mathcal{P}$ in \Cref{alg:lift,alg:Stiefel_initialization} are chosen as products of normal distributions, called with $\mathtt{randn}(N, n)$ in many programming languages. We summarize the experimental setup in \Cref{tab:experimental_setup}.

\begin{algorithm}
\caption{Initializing elements on the Stiefel manifold}\label{alg:Stiefel_initialization}
\begin{algorithmic}
    \State $A\gets{}\mathcal{P}(\mathbb{R}^{N\times{}n})$, \Comment{sample $A\in\mathbb{R}^{N\times{}n}$ from a given distribution.}
    \State $Q,R\gets\mathtt{qr}(A)$ \Comment{apply a $QR$ decomposition.}
    \State $Q\mathtt{[:,1:n]}$ \Comment{output the first $n$ columns of $Q$.}
\end{algorithmic}
\end{algorithm}

\begin{table}[tbph]
\caption{Summary of the experimental setup. The results are summarized in \Cref{tab:mnist_outcomes} and \Cref{fig:mnist_seed_band}.}\label{tab:experimental_setup}
\centering
\begin{tabular}{@{}ll}
\toprule
\textbf{Parameter} & \textbf{Value / Specification} \\
\midrule
\multicolumn{2}{l}{\textbf{Model Architecture}} \\
\hspace{1em}Model Type & Vision Transformer \\
\hspace{1em}Transformer Layers & 16 \\
\hspace{1em}Attention Mechanism & Multi-Head Attention \\
\hspace{1em}\hspace{1em}Attention Heads & 7 \\
\hspace{1em}\hspace{1em}Input Dimension & 49 ($N=49$) \\
\hspace{1em}\hspace{1em}Projection Dimension & 49/7 per head ($n=49/7=7$) \\
\hspace{1em}Feed-Forward Network & Residual layer $x\mapsto x+\sigma(Ax+b)$ (49-dim to 49-dim) \\
\hspace{1em}\hspace{1em}FFN Activation & $\sigma = \tanh$ \\
\hspace{1em}Classification Layer & Linear layer (49-dim to 10-dim) + Softmax \\
\addlinespace
\multicolumn{2}{l}{\textbf{Datasets}} \\
\hspace{1em}Datasets Used & MNIST \& Fashion-MNIST \\
\hspace{1em}Training Set Size & 60,000 images \\
\hspace{1em}Test Set Size & 10,000 images \\
\addlinespace
\multicolumn{2}{l}{\textbf{Training}} \\
\hspace{1em}Epochs & 500 \\
\hspace{1em}Batch Size & 2048 \\
\addlinespace
\multicolumn{2}{l}{\textbf{Initialization}} \\
\hspace{1em}Vector Space Weights & Glorot Uniform \cite{pmlr-v9-glorot10a} \\
\hspace{1em}Stiefel Manifold Weights & QR decomp. of random matrix normal to $Y$ (see \Cref{alg:Stiefel_initialization}) \\
\addlinespace
\multicolumn{2}{l}{\textbf{Seeds and Repetitions}} \\
\hspace{1em}Seeds & 10 fixed seeds, one per repetition, paired across configurations \\
\addlinespace
\multicolumn{2}{l}{\textbf{Software \& Hardware}} \\
\hspace{1em}Software & Julia 1.12.7 \\
\hspace{1em}Optimizer Package & \texttt{GeometricOptimizers.jl} v0.7.0 \\
\hspace{1em}Training Stack & \texttt{GMLDatasets.jl} with \texttt{GeometricMachineLearning.jl} v0.7.0 \\
\hspace{1em}GPU Backend & \texttt{CUDA.jl} 5.11.3 (CUDA 13.2) \\
\hspace{1em}Hardware & NVIDIA GeForce RTX 4090 (24 GiB) \\
\hspace{1em}Precision & Single (Float32) \\
\bottomrule
\end{tabular}
\end{table}

The training loss on MNIST with the Cayley retraction\footnote{For the standard Adam ablation we do not require the retraction as the attention-related weights $W^Q_i,$ $W^K_i$ and $W^V_i$ are in $\mathbb{R}^{49\times7}$.} for the different optimizer is reported in \Cref{transformer_statistics} as mean and sample standard deviation over \mnistSeeds{} paired seeds.

For all of these computations no hyperparameter tuning has been performed. We can see that the vision transformer without regularization, dropout or normalization is not able to learn much, as the training loss is stuck at around 1.34, which indicates a trivial prediction (see \Cref{softmax}).


\subsection{Statistical Evaluation over Ten Seeds}
\label{transformer_statistics}

\begin{figure}
    \centering
    \includegraphics[width = .8\textwidth]{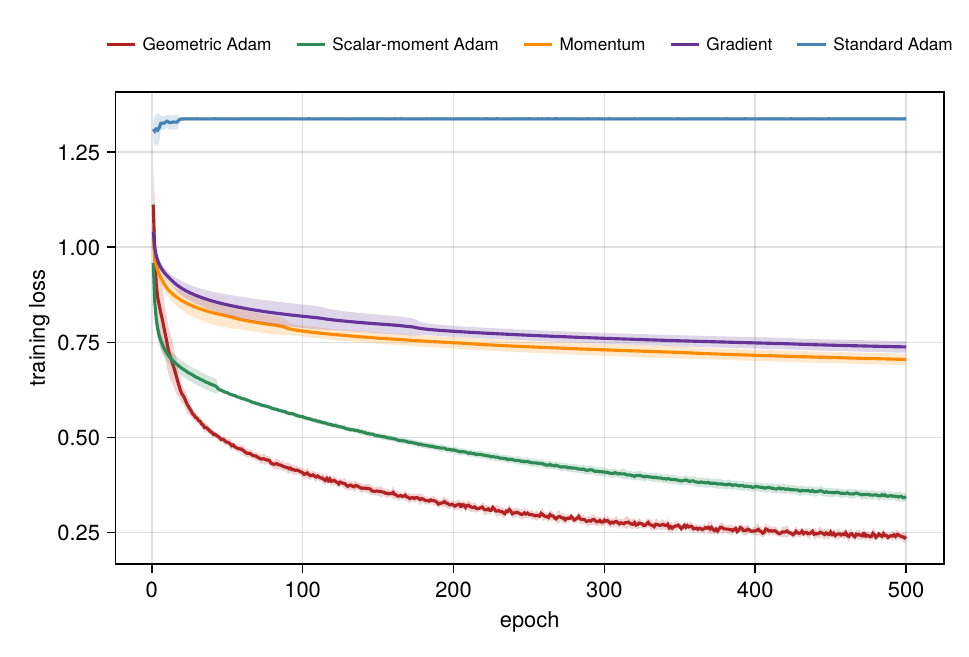}
    \caption{Training loss of the vision transformer of \Cref{fig:preprocessing} on MNIST with the Cayley retraction. Solid lines are the mean over the ten repetitions of \Cref{tab:mnist_outcomes} and the bands are one sample standard deviation.}
    \label{fig:mnist_seed_band}
\end{figure}

We train every configuration over \mnistSeeds{} seeds that are fixed in advance and report \textit{mean} and \textit{sample standard deviation} over the repetitions. The results below are the 50 runs of the MNIST data set with the Cayley retraction, \mnistEpochs{} epochs each, and are summarized in \Cref{tab:mnist_outcomes} and \Cref{fig:mnist_seed_band}.

\begin{table}[tbph]
\caption{Outcome of the vision transformer on MNIST with the Cayley retraction over 10 seeds, 500 epochs each. Entries are mean $\pm$ sample standard deviation over all repetitions. The last column is the final orthonormality drift $\|Y^TY-\mathbb{I}\|$, which is undefined for the standard Adam ablation. The first four rows constrain the projection matrices to $St(n, N)$; the scalar-moment row is the Riemannian Adam of \cite{li2020efficient}, whose second moment is one scalar per weight, and the last row is the standard Adam ablation. Hardware: NVIDIA GeForce RTX 4090, \texttt{Float32}.}
\label{tab:mnist_outcomes}
\centering
\footnotesize
\setlength{\tabcolsep}{4pt}
\begin{tabular}{@{}lrrrr@{}}
\toprule
Configuration & Final loss & Best loss & Test accuracy (\%) & $\|Y^TY-\mathbb{I}\|$ \\
\midrule
Geometric Adam & $0.240 \pm 0.014$ & $0.226 \pm 0.009$ & $86.02 \pm 0.57$ & 8.5--9.0$\,{\times}10^{-3}$ \\
Scalar-moment Adam & $0.344 \pm 0.010$ & $0.340 \pm 0.011$ & $76.20 \pm 0.75$ & 3.9--8.4$\,{\times}10^{-4}$ \\
Riemannian momentum & $0.705 \pm 0.015$ & $0.705 \pm 0.015$ & $62.73 \pm 1.77$ & 3.4--4.8$\,{\times}10^{-4}$ \\
Riemannian gradient & $0.738 \pm 0.015$ & $0.738 \pm 0.015$ & $59.11 \pm 1.88$ & 2.9--5.5$\,{\times}10^{-4}$ \\
Standard Adam & $1.337 \pm 0.005$ & $1.299 \pm 0.040$ & $10.69 \pm 0.73$ & --- \\
\botrule
\end{tabular}
\end{table}

The most important comparison is the one between the novel Adam optimizer on the Stiefel manifold and the scalar-moment baseline of \cite{li2020efficient}, since the two differ only in how the second moment is treated (compare \Cref{eq:adam_velocity,eq:scalar_adam}). Over the ten repetitions the former performs much better in terms of the training loss (see \Cref{fig:mnist_seed_band}) and test accuracy (see \Cref{tab:mnist_outcomes}): it reaches a test accuracy of \mnistGeometricAccuracy\% and the latter \mnistBaselineAccuracy\%.

\subsection{The Same Comparison on Fashion-MNIST}
\label{fashion_transformer_statistics}

\begin{figure}
    \centering
    \includegraphics[width = .8\textwidth]{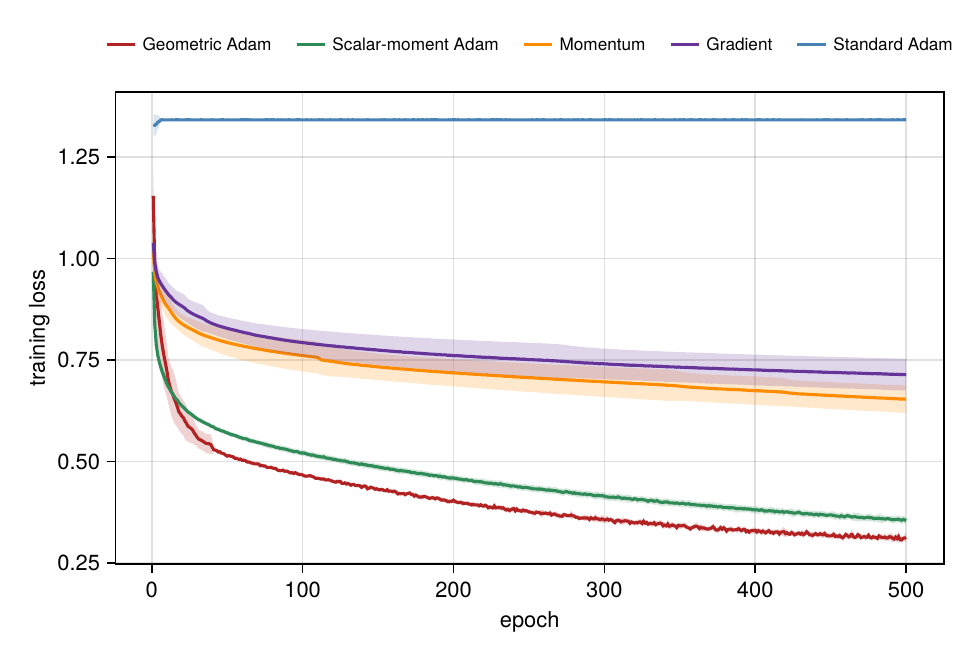}
    \caption{Training loss of the vision transformer of \Cref{fig:preprocessing} on Fashion-MNIST with the Cayley retraction. Solid lines are the mean over the ten repetitions of \Cref{tab:fashion_outcomes} and the bands are one sample standard deviation.}
    \label{fig:fashion_seed_band}
\end{figure}

We repeat the Cayley-retraction comparison on Fashion-MNIST with the same \fashionSeeds{} fixed, paired seeds and \fashionEpochs{} epochs per run. The outcomes are summarized in \Cref{tab:fashion_outcomes} and \Cref{fig:fashion_seed_band}.

\begin{table}[tbph]
\caption{Outcome of the vision transformer on Fashion-MNIST with the Cayley retraction over 10 seeds, 500 epochs each. The contents of this table are analogous to \Cref{tab:mnist_outcomes}.}
\label{tab:fashion_outcomes}
\centering
\footnotesize
\setlength{\tabcolsep}{4pt}
\begin{tabular}{@{}lrrrr@{}}
\toprule
Configuration & Final loss & Best loss & Test accuracy (\%) & $\|Y^TY-\mathbb{I}\|$ \\
\midrule
Geometric Adam & $0.313 \pm 0.012$ & $0.299 \pm 0.003$ & $78.66 \pm 0.51$ & 8.6--8.9$\,{\times}10^{-3}$ \\
Scalar-moment Adam & $0.354 \pm 0.008$ & $0.351 \pm 0.008$ & $75.35 \pm 0.62$ & 2.6--3.9$\,{\times}10^{-4}$ \\
Riemannian momentum & $0.654 \pm 0.034$ & $0.653 \pm 0.034$ & $67.43 \pm 3.63$ & 2.4--6.8$\,{\times}10^{-4}$ \\
Riemannian gradient & $0.714 \pm 0.038$ & $0.713 \pm 0.039$ & $60.98 \pm 4.66$ & 2.6--8.8$\,{\times}10^{-4}$ \\
Standard Adam & $1.342 \pm 0.000$ & $1.325 \pm 0.028$ & $10.00 \pm 0.00$ & --- \\
\botrule
\end{tabular}
\end{table}

The same comparison between the coordinate-wise and scalar second moments again favours geometric Adam: over its \fashionGeometricAccepted{} repetitions it reaches \fashionGeometricAccuracy\% test accuracy, compared with \fashionBaselineAccuracy\% for the scalar-moment baseline (\Cref{tab:fashion_outcomes}). The shared optimizer-state count in \Cref{tab:mnist_state} applies without change because the Fashion-MNIST experiment uses the same architecture. The corresponding runtime and memory measurements are reported together with the MNIST numbers in \Cref{transformer_cost}.

\subsection{Computational Cost, Runtime Overhead and Memory}
\label{transformer_cost}

The cost of optimizing on a manifold rather than on a vector space with the framework presented here is the price of two additional operations per step: mapping the Euclidean gradient to an update direction in $\mathfrak{g}^\mathrm{hor}$, and applying the retraction. Both are reported separately in \Cref{tab:mnist_cost} for MNIST and in \Cref{tab:fashion_cost} for Fashion-MNIST.

For Adam on the Stiefel manifold the gradient by automatic differentiation takes \mnistGradientMs{} ms, computing the update direction from the optimizer state takes \mnistDirectionMs{} ms and applying the Cayley retraction takes \mnistRetractionMs{} ms, so the whole geometric part is about \mnistGeometricShare\% of the optimization step. Compared to the standard Adam ablation, which trains weights of the same shape with the same automatic differentiation, the optimization step is more than $10\%$ longer. The geometric operations are therefore not the bottleneck: automatic differentiation through 16 transformer blocks at a batch size of 2048 is expensive enough that the retraction is a minority of the step.

The corresponding runtime and memory measurements on Fashion-MNIST are reported in \Cref{tab:fashion_cost} and show the same picture: for geometric Adam, the gradient by automatic differentiation takes \fashionGradientMs{} ms per step, the update direction takes \fashionDirectionMs{} ms and the Cayley retraction takes \fashionRetractionMs{} ms, so these geometric operations are about \fashionGeometricShare\% of the instrumented step, and relative to the standard Adam ablation the instrumented step is more than $10\%$ longer.

We report the \textit{memory used by the optimizer} in two different senses. The first is the peak device memory of the run, given in the last column of \Cref{tab:mnist_cost} and \Cref{tab:fashion_cost}. At the relatively big batch size of 2048, it is dominated by the automatic differentiation, it comes to about 22 GiB on both data sets\footnote{This is close to the capacity of the GPU used (24 GiB), and the batch size has been chosen so that this is the case.}. The second is the footprint of the optimizer state itself, which is a structural property of the representation and is counted in \Cref{tab:mnist_state}. An element of $\mathfrak{g}^\mathrm{hor}$ for $St(7,49)$ is stored as a skew-symmetric $7\times7$ block and a dense $42\times7$ block, i.e.\ \mnistHorizontalEntries{} entries (see \Cref{eq:g_hor}), against \mnistAmbientEntries{} for the Euclidean representative. In addition, every GO optimizer state stores a \texttt{GlobalSection}: for each Stiefel weight this is the full $49\times49$ $[Y,\lambda]$ matrix, i.e.\ \mnistGlobalSectionEntries{} entries, while ordinary parameters contribute sections matching their array shapes. Adam carries two moment representations per weight. Over the 16 layers of this network, including the global sections, that gives \mnistGeometricStateMiB{} MiB of optimizer state, against \mnistAblationStateMiB{} MiB for the standard Adam ablation and \mnistBaselineStateMiB{} MiB for the scalar-moment baseline.

\begin{table}[tbph]
\caption{Computational cost of one training run on MNIST with the Cayley retraction, on an NVIDIA GeForce RTX 4090 in \texttt{Float32}. The cost of the optimizer step is split into the automatic differentiation part, the computation of the \textit{final velocity} (see \Cref{alg:general_man}) from the optimizer state, and the application of the retraction. The ``unattributed'' column is the total wall-clock of the run divided by its steps, minus those three; it collects data movement, host synchronization and the per-epoch evaluation of the test accuracy. ``Geometric share'' is direction plus retraction as a fraction of the instrumented step. The time per epoch is the median over all repetitions. Peak device memory is the maximum over the run.}
\label{tab:mnist_cost}
\centering
\footnotesize
\setlength{\tabcolsep}{3pt}
\begin{tabular}{@{}lrrrrrrr@{}}
\toprule
 & \multicolumn{4}{c}{Time per step (ms)} &  &  &  \\
\cmidrule(lr){2-5}
Configuration & Gradient & Direction & Retraction & Unattri- & Geometric & Time per & Peak device \\
 & (AD) &  &  & buted & share (\%) & epoch (s) & mem.\ (GiB) \\
\midrule
Geometric Adam & 118.4 & 4.85 & 20.83 & 120.4 & 17.8 & 7.67 & 21.8 \\
Scalar-moment Adam & 115.8 & 5.33 & 36.15 & 44.9 & 25.8 & 5.63 & 22.2 \\
Riemannian momentum & 121.5 & 4.33 & 41.39 & 126.6 & 25.6 & 7.91 & 23.5 \\
Riemannian gradient & 131.2 & 5.37 & 32.95 & 136.3 & 22.5 & 8.73 & 23.4 \\
Standard Adam & 119.7 & 9.26 & 0.30 & 118.4 & 7.4 & 7.21 & 22.6 \\
\botrule
\end{tabular}
\end{table}

\begin{table}[tbph]
\caption{Computational cost of one training run on Fashion-MNIST with the Cayley retraction, on an NVIDIA GeForce RTX 4090 in \texttt{Float32}. The contents of this table are analogous to \Cref{tab:mnist_cost}.}
\label{tab:fashion_cost}
\centering
\footnotesize
\setlength{\tabcolsep}{3pt}
\begin{tabular}{@{}lrrrrrrr@{}}
\toprule
 & \multicolumn{4}{c}{Time per step (ms)} &  &  &  \\
\cmidrule(lr){2-5}
Configuration & Gradient & Direction & Retraction & Unattri- & Geometric & Time per & Peak device \\
 & (AD) &  &  & buted & share (\%) & epoch (s) & mem.\ (GiB) \\
\midrule
Geometric Adam & 117.0 & 4.86 & 20.82 & 121.7 & 18.0 & 7.67 & 22.2 \\
Scalar-moment Adam & 115.2 & 4.94 & 20.67 & 44.3 & 18.2 & 5.37 & 22.2 \\
Riemannian momentum & 119.6 & 4.03 & 24.44 & 123.5 & 19.2 & 7.87 & 23.5 \\
Riemannian gradient & 118.6 & 5.04 & 24.49 & 124.8 & 19.9 & 7.90 & 23.4 \\
Standard Adam & 119.0 & 9.88 & 0.30 & 119.1 & 7.9 & 7.22 & 22.7 \\
\botrule
\end{tabular}
\end{table}

\begin{table}[tbph]
\caption{Counted optimizer-state footprint of the vision transformer of \Cref{fig:preprocessing}. The network has $336$ attention projections on $St(7, 49)$ and $39690$ further scalar parameters. An element of $\mathfrak{g}^\mathrm{hor}$ is stored as a skew-symmetric $7\times7$ block and a dense $42\times7$ block (see \Cref{eq:g_hor}), i.e.\ $315$ entries against $343$ for the ambient representative and $1176$ for a naive element of $\mathfrak{g}=\mathfrak{so}(49)$. Every GO optimizer state also stores a \texttt{GlobalSection}; for a Stiefel weight this is the full $[Y,\lambda]$ matrix of size $49\times49$, i.e.\ $2401$ entries, while ordinary parameters contribute sections matching their array shapes. Sizes are for \texttt{Float32}. This is a structural count and not a measurement.}
\label{tab:mnist_state}
\centering
\begin{tabular}{@{}lrrr@{}}
\toprule
Configuration & Entries per & Total state & State \\
 & Stiefel weight & (entries) & (MiB) \\
\midrule
Geometric Adam & 3031 & 1137486 & 4.34 \\
Scalar-moment Adam & 2717 & 1031982 & 3.94 \\
Riemannian momentum & 2716 & 991956 & 3.78 \\
Riemannian gradient & 2401 & 846426 & 3.23 \\
Standard Adam & 1029 & 464814 & 1.77 \\
\botrule
\end{tabular}
\end{table}

\subsection{Practical Limitations and Unexplained Observations}
\label{transformer_limitations}

Here we list a few observations of the experiments that remain unexplained and may constitute future work.

\textit{The drift off the manifold is larger for the proposed optimizer than for any other geometric configuration.} The Cayley retraction maps onto the Stiefel manifold in exact arithmetic, so the drift in the last column of \Cref{tab:mnist_outcomes} is accumulated rounding in single precision over the \mnistEpochs{} epochs. The Adam runs on the Stiefel manifold finish between \mnistDriftMin{} and \mnistDriftMax, significantly higher than the drift of the scalar-moment baseline, the momentum optimizer and the gradient optimizer, which all stay below $10^{-3}$. This is probably a fact about the method: a coordinate-wise second moment produces a longer step than a single scalar does, and a longer step leaves more of the manifold behind it. Mitigating this issue may be a subject of future research.

\textit{The Riemannian gradient and the Riemannian momentum optimizer seem to incur more overhead than Adam.} The \textit{Riemannian gradient} and \textit{Riemannian momentum} optimizers in \Cref{tab:mnist_cost,tab:fashion_cost} seem unexpectedly more expensive than the \textit{geometric Adam}, even though they perform fewer computations. Why and how this can be the case may be another direction for future research.

\section{Numerical Example: Symplectic Model Reduction of the Pendulum}
\label{pendulum_sae}

As a second application we consider training a \textit{symplectic autoencoder} (SAE) \cite{brantner2023symplectic} to reduce a higher-dimensional system to a lower-dimensional one. Different from the transformer in the previous example, optimization on manifolds is not only a tool to increase efficiency in training but is required by construction: the SAE would not maintain its symplecticity property if we do not optimize on manifolds, so manifold optimization is essential here. We stress that this example is academic in nature: the pendulum is a classical toy problem, and the example serves solely to demonstrate the efficacy of the optimizer developed in this work.

The network used here is the symplectic autoencoder introduced in \cite{brantner2023symplectic}. It consists of an \textit{encoder} $\Psi^e:\mathbb{R}^{2N}\to\mathbb{R}^{2n}$ and a \textit{decoder} $\Psi^d:\mathbb{R}^{2n}\to\mathbb{R}^{2N}$ ($n\ll{}N$) that map the data to a low-dimensional space and back while preserving the canonical symplectic structure. The crucial component in SAEs are \textit{proper symplectic decomposition} (PSD) layers, which can change dimension while preserving structure. The PSD layers have the form
\begin{equation}
    A = \begin{pmatrix} \Phi & 0 \\ 0 & \Phi \end{pmatrix}, \qquad \Phi\in{}St(n, N),
    \label{eq:psd_layer}
\end{equation}
i.e. their weights lie on the Stiefel manifold. Training the SAE therefore amounts to minimizing the reconstruction error on a product of Stiefel manifolds and vector spaces: the symplecticity of the network is preserved by construction only if the Stiefel-constrained weights are updated by manifold optimization, which is a direct application of the optimizer developed in this work. In the present example we have $2N=4$ and $2n=2$.

The problem the SAE is trained on is the classical pendulum. In \textit{angular coordinates} the state of the system is the pair $(\theta,\dot\theta)$, where $\theta$ is the angular coordinate and $\dot\theta$ its time derivative; the dynamics is the autonomous system $\ddot\theta = \sin\theta$. The training data in \texttt{GMLDatasets.jl} \cite{brantner2026gmldatasets} comprises 100 trajectories of this system whose initial conditions form a $10\times10$ grid on $[0,2\pi]\times[-2,2]$. Each trajectory is generated on $t\in[0,10]$ with step size $0.1$; the choice of initial conditions covers both the oscillatory and the rotational solutions of the system (i.e. they cross the separatrix).

Every state is then lifted from the angular coordinates to the full Euclidean coordinates (also see \Cref{fig:pendulum_sae_workflow})
\begin{equation}
    q=(\sin\theta,\cos\theta),\qquad
    p=(\dot\theta\cos\theta,-\dot\theta\sin\theta).
    \label{eq:pendulum_lift}
\end{equation}
The SAE is trained on these four-dimensional data to learn a two-dimensional representation of it, i.e. reduced coordinates in
which the dynamics of the system can be described. The overall workflow is summarized in
\Cref{fig:pendulum_sae_workflow}.

\begin{figure}
    \centering
    \begin{tikzpicture}[
    box/.style={draw, very thick, rounded corners, minimum width=3.6cm, minimum height=1.9cm},
    arrow/.style={-{Latex[length=.7cm, width=.35cm]}, ultra thick},
]

\node[box, fill=mblue!25] (angular) at (0,0) {\begin{tabular}{c}
    Data in angular\\[.2ex] coordinates $(\theta,\,\dot\theta)$
    \end{tabular}};

\node[box, fill=mpurple!25, xshift=5.4cm] (full) {\begin{tabular}{c}
    Lift to full\\[.2ex] coordinates $(q,p)\in\mathbb{R}^4$
    \end{tabular}};

\node[box, fill=morange!25, xshift=10.8cm] (sae) {\begin{tabular}{c}
    SAE\\[.2ex] $\mathbb{R}^4\overset{\Psi^e}{\to}\mathbb{R}^2\overset{\Psi^d}{\to}\mathbb{R}^4$
    \end{tabular}};

\node[box, fill=mgreen!25, xshift=16.2cm] (latent) {\begin{tabular}{c}
    Learned\\[.2ex] coordinates in $\mathbb{R}^2$
    \end{tabular}};

\draw[arrow] (angular.east) -- node[above, font=\small] {lift} (full.west);
\draw[arrow] (full.east) -- (sae.west);
\draw[arrow] (sae.east) -- node[above, font=\small] {reduce} (latent.west);
\draw[arrow, dashed, mred] (sae.north) to[bend left=-30] node[below, font=\small] {reconstruct} (full.north);

\end{tikzpicture}
    \caption{Workflow of the pendulum example: the data are generated in the angular coordinates of the pendulum, lifted to the full four-dimensional coordinates, reduced with the symplectic autoencoder, and the result is a learned two-dimensional set of coordinates that is different from $(\theta,\dot{\theta})$.}
    \label{fig:pendulum_sae_workflow}
\end{figure}

The setup mirrors the MNIST experiment above: its architecture and optimization parameters are
fixed before the full run, while initialization and batch order vary with ten fixed seeds. The loss
is the Euclidean reconstruction error, i.e. the sum over data points $z$ of $\|z - \Psi^d\circ\Psi^e(z)\|_2^2$. The relevant settings are
summarized in \Cref{tab:pendulum_sae_setup}.

\begin{table}[tbph]
\caption{Setup of the pendulum symplectic-autoencoder experiment. The results are summarized in \Cref{tab:pendulum_outcomes} and \Cref{fig:pendulum_seed_band}.}\label{tab:pendulum_sae_setup}
\centering
\begin{tabular}{@{}ll}
\toprule
\textbf{Parameter} & \textbf{Value / Specification} \\
\midrule
\multicolumn{2}{l}{\textbf{Model Architecture}} \\
\hspace{1em}Model Type & Symplectic Autoencoder (SAE) \\
\hspace{1em}Dimensions & $\mathbb{R}^4\rightarrow\mathbb{R}^2\rightarrow\mathbb{R}^4$ \\
\hspace{1em}Encoder / decoder & 2 blocks each; 10 / 20 layers per block \\
\hspace{1em}Decoder output / upscale & 10 / 20 \\
\addlinespace
\multicolumn{2}{l}{\textbf{Datasets}} \\
\hspace{1em}Datasets Used & Pendulum (generated) \\
\hspace{1em}Training Set Size & 100 trajectories, 101 time points each \\
\addlinespace
\multicolumn{2}{l}{\textbf{Training}} \\
\hspace{1em}Epochs & \pendulumEpochs{} \\
\hspace{1em}Batch Size & 256 \\
\addlinespace
\multicolumn{2}{l}{\textbf{Initialization}} \\
\hspace{1em}Vector Space Weights & Glorot Uniform \cite{pmlr-v9-glorot10a} \\
\hspace{1em}Stiefel Manifold Weights & QR decomp. of random matrix (see \Cref{alg:Stiefel_initialization}) \\
\addlinespace
\multicolumn{2}{l}{\textbf{Seeds and Repetitions}} \\
\hspace{1em}Seeds & \pendulumSeeds{} fixed seeds, one per repetition, paired across configurations \\
\addlinespace
\multicolumn{2}{l}{\textbf{Software \& Hardware}} \\
\hspace{1em}Software & Julia 1.12.7 \\
\hspace{1em}Optimizer Package & \texttt{GeometricOptimizers.jl} v0.7.0 \\
\hspace{1em}Training Stack & \texttt{GMLDatasets.jl} with \texttt{GeometricMachineLearning.jl} v0.7.0 \\
\hspace{1em}GPU Backend & \texttt{CUDA.jl} 5.11.3 (CUDA 13.2) \\
\hspace{1em}Hardware & NVIDIA GeForce RTX 4090 (24 GiB) \\
\hspace{1em}Precision & Single (Float32) \\
\bottomrule
\end{tabular}
\end{table}

The $\mathcal{O}$-parameters $\Xi$ (see \Cref{alg:general_man}) for the four configurations are set to the following values:

\begin{table}[tbph]
    \centering%
    \begin{tabular}{|lllll|}
        \toprule%
        Gradient optimizer:& $\eta = 10^{-4}$. &                  &                   & \\
        Momentum optimizer:& $\eta = 10^{-4}$,  & $\alpha = 0.5$.  &                   & \\
        Adam optimizer:    & $\eta = 10^{-4}$, & $\beta_1 = 0.9$, & $\beta_2 = 0.99$, & $\delta=10^{-8}$.\\
        Scalar-moment Adam:& $\eta = 10^{-4}$, & $\beta_1 = 0.9$, & $\beta_2 = 0.99$, & $\delta=10^{-8}$.\\
        \botrule%
    \end{tabular}
\end{table}

\subsection{Statistical Evaluation over Ten Seeds}
\label{pendulum_sae_statistics}

\begin{figure}
    \centering
    \includegraphics[width=.8\textwidth]{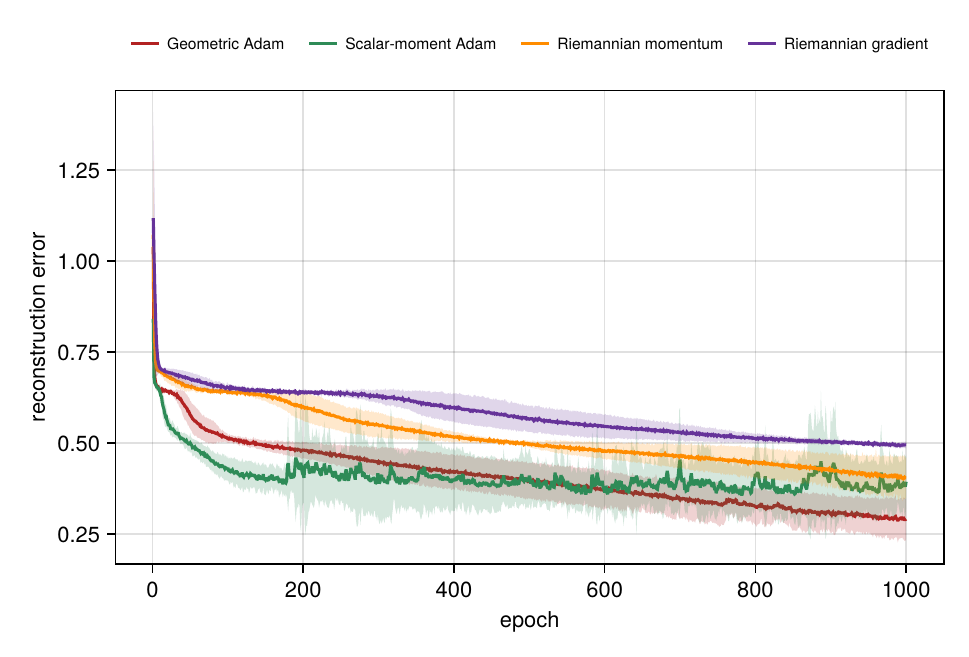}
    \caption{Reconstruction error of the pendulum symplectic autoencoder. Solid lines are the mean over the \pendulumSeeds{} paired seeds and the bands are one sample standard deviation.}
    \label{fig:pendulum_seed_band}
\end{figure}

The outcomes of the pendulum trainings are reported in \Cref{tab:pendulum_outcomes}; the corresponding
mean reconstruction-error curves and one sample-standard-deviation bands are shown in
\Cref{fig:pendulum_seed_band}.

\begin{table}[tbph]
\caption{Outcome of the pendulum symplectic-autoencoder experiment over 10 paired seeds and 1000 epochs each. Entries are mean $\pm$ sample standard deviation over all repetitions. The scalar-moment row is the Riemannian Adam baseline of \cite{li2020efficient}; it differs from geometric Adam only in its scalar rather than coordinate-wise second moment. The times are complete training-run wall-clock times on an NVIDIA GeForce RTX 4090 in \texttt{Float32}.}
\label{tab:pendulum_outcomes}
\centering
\footnotesize
\setlength{\tabcolsep}{4pt}
\begin{tabular}{@{}lrrr@{}}
\toprule
Configuration & Final reconstruction & Best reconstruction & Training time \\
 & error & error & (min) \\
\midrule
Geometric Adam & $0.292 \pm 0.055$ & $0.281 \pm 0.059$ & $17.84 \pm 0.05$ \\
Scalar-moment Adam & $0.394 \pm 0.072$ & $0.289 \pm 0.050$ & $17.91 \pm 0.05$ \\
Riemannian momentum & $0.405 \pm 0.057$ & $0.393 \pm 0.061$ & $13.63 \pm 0.06$ \\
Riemannian gradient & $0.495 \pm 0.009$ & $0.488 \pm 0.007$ & $12.42 \pm 0.06$ \\
\botrule
\end{tabular}
\end{table}

The primary comparison is geometric Adam against the scalar-moment Riemannian-Adam baseline
of \cite{li2020efficient}: these configurations share the learning rate, retraction and transport,
and differ only in whether their second moment is coordinate-wise or scalar. After
\pendulumEpochs{} epochs, geometric Adam reaches a final reconstruction error of
\pendulumGeometricFinal, compared with \pendulumBaselineFinal{} for the scalar-moment baseline
(\Cref{tab:pendulum_outcomes}). This is a \pendulumFinalReduction\% lower mean final error. As in the MNIST and Fashion-MNIST examples, the gradient and momentum rows provide further geometric reference points. Also note how in \Cref{fig:pendulum_seed_band} the losses for the geometric Adam and the scalar-moment baseline are qualitatively different: the loss for geometric Adam reaches a lower level and exhibits almost no oscillations.


\section{Conclusion and Outlook}

A generalization of Adam to homogeneous spaces (such as the Stiefel manifold) has been presented. Unlike other \textit{start-of-the-art} manifold optimizers \cite{li2020efficient, kong2022momentum} we manage to fully generalize the Adam optimizer without relying on projections or a simplified cache. We do so by exploiting the key structural property of these spaces: the existence of a global tangent space representation (called ``Lie subspace'' in \cite{o1983semi}), which is the generalization of the Lie algebra for Lie groups. This allows all of Adam's operations to be performed in a consistent vector space, similarly as was done previously for Lie groups in e.g. \cite{lezcano2019cheap}.

The numerical experiments demonstrated that the resulting optimizers can simplify the training of transformer neural networks: on MNIST and Fashion-MNIST, the proposed optimizer outperformed the baseline \cite{li2020efficient} and managed to efficiently train a deep transformer without dropout, layer normalization or regularization. We further showed that the introduction of geometric tools only slightly increases computational cost compared to the standard Adam ablation.

In addition, we demonstrated the efficacy of the new optimizer on the pendulum symplectic autoencoder \cite{brantner2023symplectic}, where the network's symplecticity is preserved only if the Stiefel-constrained weights are updated by manifold optimization, in contrast to the transformer example, where manifold optimization is a tool to increase efficiency. We again observed as similar (if not greater) increases in efficiency and performance as in the MNIST/Fashion-MNIST experiments.

This makes the new optimizers suitable for many applications. They can (i) facilitate training for relatively big transformers and (ii) enable training networks that rely on a specific structure, as the symplectic autoencoder example showed, where the structure is preserved by construction \cite{brantner2023symplectic}. Regarding point (i), enforcing hard geometric constraints can be a more principled and effective alternative to heuristic techniques like layer normalization, dropout, or regularization and simplifying the training of complex models like transformers. This work also solves the challenge posed by \cite{lezcano2019cheap} as it ``generalizes [previous optimizers] to homogeneous Riemannian manifolds, like the Stiefel manifold.''


Future work will focus on both further implementation and theoretical aspects. In terms of implementation, we aim to use the presented framework to generalize the BFGS optimizer and others \cite{wright2006numerical}, as well as apply our method to a wider range of problems. From a theoretical point of view we will further investigate possible isotropy choices as discussed in \cite{krogstad2003low,munthe2016integrators}, the source of the unexplained timing differences, and ways to control the orthogonality drift. Another promising direction is an adaptation of the ideas behind AdamW \cite{loshchilov2017decoupled} to homogeneous spaces: as the Euclidean $L^2$ penalty is constant on compact manifolds such as the Stiefel manifold, this would have to be based either on a norm that is not constant on the manifold or on a decay of the global tangent space coordinates instead of the weights themselves.

\section*{Acknowledgements}
The author wants to express his deep gratitude to Michael Kraus, who co-developes the \texttt{GeometricOptimizers.jl} library with him and without whom the author would not have been able to obtain the newest batch of experimental results. The author further wants to thank Tobias Blickhan for initial discussion and help with implementing a first version of \Cref{alg:exp_it}.

\phantomsection
\addcontentsline{toc}{section}{References}
\bibliographystyle{plainnat}
\bibliography{optimizer}

\newpage
\appendix
\addappheadtotoc

\section{Induced Isomorphism}
\label{induced_isomorphism}

The mapping $\Omega$ in equation \Cref{eq:omega} can be easily shown to establish the isomorphism $T_Y\mathcal{M}\overset{\sim}{\to}\mathfrak{g}^{\mathrm{hor},Y}$: 
\begin{multline}
    \Omega(V_Y)Y = \bigg(\mathbb{I} - \frac{1}{2}YY^T\bigg)V_Y - \frac{1}{2}Y(V_Y)^TY = \\
    \bigg(\mathbb{I} - \frac{1}{2}YY^T\bigg)V_Y + \frac{1}{2}YY^TV_Y = V_Y,
\end{multline}
where it was used that $V_Y\in{}T_Y\mathcal{M}$, i.e. $(V_Y)^TY = -Y^TV_Y$.

\section{Computing the Lift $Y\mapsto\lambda(Y)$}
\label{global_section}

\Cref{alg:lift} uses a $QR$ decomposition, or more precisely, Householder reflections. These are implemented in most numerical linear algebra libraries. 

In essence, Householder reflections take as input an $N\times{}M$ matrix and transform it to an upper-triangular one by a series of $M$ rotations. These rotations are stored in the $Q$ matrix of the $QR$ decomposition.\footnote{$R$ is the upper-triangular matrix resulting from the transformations. A detailed description of Householder reflections is given in e.g. \cite{mezzadri2006generate}.}

The computation of the section $\Lambda$ is shown in \Cref{alg:lift}. The following is needed to ensure $\lambda$ is in fact a lift to $G=SO(N)$:
\begin{proposition}
    Let $Y\in\mathbb{R}^{N\times{}n}$ be a matrix whose columns are orthonormal, $A\in\mathbb{R}^{N\times{}(N-n)}$ be such that $Y^TA = \mathbb{O}_n$ and $QR = A$ its decomposition. Then $Q^TY = \mathbb{O}_n$.
\end{proposition}
\begin{proof}
    Write $Q = [q_1, \ldots, q_{N-n}, \ldots]$. The special structure of $R$ (i.e. $[R]_{ij} =: r_{ij} = 0$ for $i>j$) means that the first column of $A =: [a_1, \ldots, a_{N-n}]$ is a linear combination of $\{q_1\}$, the second column is a linear combination of $\{q_1,q_2\}$ and so forth. 
    But this in turn means that every $q_i$ ($i=1,\ldots,(N-n$)) can be constructed with columns of $A$. Thus $Q$ and $A$ span the same vector space, and this vector space is orthogonal to $Y$ by assumption. 
\end{proof}

It should be noted that this specific part of the algorithm is also extendable to, for example, the symplectic Stiefel manifold, as there also exists a Householder routine for this \cite{salam2008symplectic}. Other $QR$ decompositions may also be used for this step.

\section{Computing Exponentials}
\label{computing_the_exponential}

As was already discussed by other authors \cite{bendokat2021real, celledoni2000approximating}, computing the matrix exponential to solve the geodesic in \Cref{th:geodesic} for a matrix manifold with large dimension $N$ and small dimension $n$ only requires computing a matrix exponential of a $2n\times2n$-dimensional matrix.

The implementation of our algorithm uses the following (also see \cite[proposition 3]{celledoni2000approximating} and \cite[proposition 3.8]{bendokat2021real}):
The elements of $\mathfrak{g}^\mathrm{hor}$ have a special block structure (see equation \Cref{eq:g_hor}) that allows each matrix to be written in the following form:

\begin{equation}
    \mathfrak{g}^\mathrm{hor} = \left\{ \begin{bmatrix}  \frac{1}{2}A & \mathbb{I}_n \\ B & \mathbb{O} \end{bmatrix} \begin{bmatrix}  \mathbb{I}_n & \mathbb{O}^T \\ \frac{1}{2}A & -B^T  \end{bmatrix} : A\in\mathbb{R}^{n\times{}n}\text{ skew-sym, }B\in\mathbb{R}^{N\times{}n}\text{ arbitrary}  \right\}.
    \label{eq:g_sparse}
\end{equation}

The computation of the geodesic can be performed cheaply by recognizing the following (the two block matrices in equation \Cref{eq:g_sparse} will be called $B'$ and $B''^T$ with $B', B''\in\mathbb{R}^{N\times2n}$, so that $\bar{B} = B'B''^T$). Since $(B'B''^T)^k = B'(B''^TB')^{k-1}B''^T$ for $k\geq1$, substituting into the exponential series gives the exact identity
\begin{equation}
    \exp(B'B''^T) = \mathbb{I}_N + B'\mathfrak{A}(X)B''^T, \qquad \mathfrak{A}(X) := \sum_{k=1}^\infty\frac{1}{k!}X^{k-1}, \qquad X := B''^TB',
    \label{eq:low_rank_identity}
\end{equation}
so the whole computation reduces to evaluating one matrix function on the $2n\times2n$ \textit{reduced matrix} $X$.

The obvious approach to computing $\mathfrak{A}(X)$ would be to sum the series until the norm of an individual term falls below a given threshold (e.g. machine precision); this amounts to Taylor-expanding the expression and then truncating. A property of $X$ makes this approach impractical however: its nonzero eigenvalues are the nonzero --- and purely imaginary --- eigenvalues of the normal matrix $\bar{B}$, since $B'B''^T$ and $B''^TB'$ have the same nonzero spectrum \cite[section 1.3]{horn2012matrix}, so its spectral radius is $\rho(X) = ||\bar{B}||_2$; but multiplying out the two factors of \Cref{eq:g_sparse} gives $X = \left[\begin{smallmatrix} \frac{1}{2}A & \mathbb{I}_n \\ \frac{1}{4}A^2 - B^TB & \frac{1}{2}A\end{smallmatrix}\right]$, whose lower-left block is \textit{quadratic} in the entries of $\bar{B}$ where every other block is linear, so $||X||$ grows like $||\bar{B}||_2^2$. Hence $\rho(X)/||X||_2 \approx ||\bar{B}||_2^{-1}$. The powers of this matrix are sized by its norm while the sum they converge to is sized by its spectrum \cite{trefethen2005spectra}, so the partial sums $S_m := \sum_{k=1}^mX^{k-1}/k!$ rise to a peak far above $\mathfrak{A}(X)$ and cancel back down to it.

The remedy is to sum at a scaled argument and transport the result back exactly. For $\exp$ this is the classical scaling-and-squaring method \cite{higham2005scaling, almohy2010new}; for $\mathfrak{A}$ the recovery step is a \textit{modified} squaring \cite{skaflestad2009scaling}, coming from the following identity: combining $\exp(2Y) - \mathbb{I} = (\exp(Y) - \mathbb{I})(\exp(Y) + \mathbb{I})$ with $\exp(Y) = \mathbb{I} + Y\mathfrak{A}(Y)$ gives
\begin{equation}
    \mathfrak{A}(2Y) = \mathfrak{A}(Y)\tfrac{1}{2}\left(\exp(Y) + \mathbb{I}\right) = \mathfrak{A}(Y) + \tfrac{1}{2}Y\mathfrak{A}(Y)^2,
    \label{eq:doubling_identity}
\end{equation}
an identity of power series, so it holds at singular $Y$ as well. Setting $W_j := \mathfrak{A}(X/2^j)/2^j$ and applying it at $Y = X/2^j$ turns it into $W_{j-1} = 2W_j + W_jXW_j$. This yields \Cref{alg:exp_it}.

\begin{algorithm}
\caption{Evaluation of $\mathfrak{A}(X)$ for $X = B''^TB'$ by scaling and modified squaring. Here $\varepsilon$ denotes machine precision and $\theta$ the scaling threshold ($\theta = 1/2$ by default).}\label{alg:exp_it}
\begin{algorithmic}
    \State $s \gets \max\left(0, \lceil\log_2(||X||_1/\theta)\rceil\right)$, $Y\gets{}X/2^s$, \Comment{so that $||Y||_1\leq\theta$}
    \State $\mathtt{output}\gets \mathbb{I}$, $\mathtt{product} \gets \mathbb{I}$, $t\gets1$,
    \While{$||\mathtt{product}|| > \varepsilon$}
    \State $t\gets{}t+1$, $\mathtt{product} \gets \frac{1}{t}\mathtt{product}\cdot{}Y$, $\mathtt{output} \gets \mathtt{output} + \mathtt{product}$
    \EndWhile
    \State $W \gets \mathtt{output}/2^s$ \Comment{$W = W_s = \mathfrak{A}(X/2^s)/2^s$}
    \For{$j = s, s-1, \ldots, 1$} \Comment{undo the halvings, with the original $X$}
    \State $W \gets 2W + WXW$
    \EndFor
    \State \Return $W$ \Comment{$W = W_0 = \mathfrak{A}(X)$}
\end{algorithmic}
\end{algorithm}

Only the summation is an approximation: the division by $2^s$ is part of the definition of $W_s$ and \Cref{eq:doubling_identity} is exact, so the recovery contributes no truncation error.

The retraction hence takes the form:
\begin{equation}
    \begin{aligned}
    \mathcal{R}^\mathrm{geo}_Y(\Delta) & \equiv \lambda(Y)\mathtt{retraction}(B) \\ & = Y + \lambda(Y)\begin{bmatrix} \frac{1}{2}A & \mathbb{I}_n \\ B & \mathbb{O} \end{bmatrix} \mathfrak{A}\left(\begin{bmatrix}  \frac{1}{2}A & \mathbb{I}_n \\ \frac{1}{4}A^2 - B^TB & \frac{1}{2}A \end{bmatrix}\right)\begin{bmatrix} \mathbb{I}_n \\ \frac{1}{2}A  \end{bmatrix},
    \end{aligned}
\end{equation}
where the matrix exponential is now only computed for a \textit{small} $2n\times2n$ matrix with \Cref{alg:exp_it}.

\section{The Cayley Retraction}
\label{cayley}

Besides the geodesic retraction, the perhaps most common example of a retraction for matrix manifolds is the \emph{Cayley retraction}. It is a retraction for many matrix Lie groups \cite{absil2008optimization,bendokat2021real}. For $V\in{}T_\mathbb{I}G\equiv\mathfrak{g} = \mathfrak{so}(N)$ it is defined as

\begin{equation}
\mathrm{Cayley}(V) = \left(\mathbb{I} - \frac{1}{2}V\right)^{-1}\left(\mathbb{I} +\frac{1}{2}V\right).
\end{equation}

%
%
%
%

We can also use the Cayley retraction at a different point than the identity $\mathbb{I}.$ For this consider $\bar{A}\in{}SO(N)$ and $\bar{B}\in{}T_{\bar{A}}SO(N) = \{\bar{B}\in\mathbb{R}^{N\times{}N}: \bar{A}^T\bar{B} + \bar{B}^T\bar{A} = \mathbb{O}\}$. We then have $\bar{A}^T\bar{B}\in\mathfrak{so}(N)$ and 
\begin{equation}
\overline{\mathrm{Cayley}}: T_{\bar{A}}SO(N) \to SO(N), \bar{B} \mapsto \bar{A}\mathrm{Cayley}(\bar{A}^T\bar{B}),
\end{equation}

is a retraction $\forall{}\bar{A}\in{}SO(N)$.

Similar to the \textit{geodesic retraction} (see \Cref{computing_the_exponential}) we also leverage the decomposition of $\bar{B} = B'(B'')^T\in\mathfrak{g}^\mathrm{hor}$ (see \Cref{eq:g_sparse}) for the Cayley retraction. Here we do this through the \emph{Sherman-Morrison-Woodbury formula} \cite{wright2006numerical}:

\begin{equation}
(\mathbb{I}_{N} - \frac{1}{2}B'(B'')^T)^{-1} = \mathbb{I}_N + \frac{1}{2}B'(\mathbb{I}_{2n} - \frac{1}{2}(B'')^TB')^{-1}(B'')^T
\end{equation}

So what we have to compute the inverse of is:

\begin{equation}
\begin{split}
\mathbb{I}_{2n} - \frac{1}{2}\begin{bmatrix}  \mathbb{I} & \mathbb{O}^T \\ \frac{1}{2}A & -B^T  \end{bmatrix}\begin{bmatrix}  \frac{1}{2}A & \mathbb{I}_n \\ B & \mathbb{O} \end{bmatrix} = 
\begin{bmatrix}  \mathbb{I}_n - \frac{1}{4}A & - \frac{1}{2}\mathbb{I}_n \\ \frac{1}{2}B^TB - \frac{1}{8}A^2 & \mathbb{I}_n - \frac{1}{4}A  \end{bmatrix},
\end{split}
\end{equation}

where $A$ and $B$ describe an element of $\mathfrak{g}^\mathrm{hor}$ (see \Cref{eq:g_hor}). By leveraging the sparse structure of the matrices in $\mathfrak{g}^\mathrm{hor}$ we arrive at the following expression for the Cayley retraction (similar to the case of the geodesic retraction):

\begin{equation}
\mathrm{Cayley}(\bar{B}) = \mathbb{I}_N + B' \left(\mathbb{I}_{2n} - \frac{1}{2} (B'')^T B'\right)^{-1} (B'')^T.
\end{equation}

Similar to the geodesic retraction, here the expensive operation (the matrix inverse in this case) is only computed for a small $2n\times2n$ matrix.

\section{Parallel Transport}
\label{parallel_transport}

Here we discuss \textit{parallel transport} of the cache along the curve defined by retractions in our algorithm. The concept of \emph{parallel transport along a geodesic} $\gamma:[0, T]\to\mathcal{M}$ describes moving a tangent vector from $T_{\gamma(0)}\mathcal{M}$ to $T_{\gamma(t)}\mathcal{M}$ such that its orientation with respect to the geodesic is preserved.

A precise definition of parallel transport needs a notion of a \emph{connection}. We refer to \cite{lang2012fundamentals,bishop1980tensor} for a detailed discussion of connections. Here we simply recall theoretical results about the Levi-Civita connection from \cite{edelman1998geometry} and the ``canonical invariant connection of second kind'' \cite{nomizu1954invariant,schlarb2024covariant}. We first state the equations for parallel transport along the Levi-Civita connection \cite[Theorem 5.11.1]{bishop1980tensor} (this connection can be seen as the \textit{natural one} for a given Riemannian metric):

\begin{proposition}
Given two elements $B^A_1, B^A_2\in{}T_AG$ the parallel transport of $B^A_2$ along the geodesic of $B^A_1$ with respect to the Levi-Civita connection is given by

\begin{equation}
    \Pi_{A\to\gamma_{B^A_1}(t)}B_2^A = A\exp(\frac{t}{2}B_1)B_2\exp(\frac{t}{2}B_1),
\end{equation}

where $B_i := A^{-1}B^A_i.$
\end{proposition}

For a more detailed discussion see e.g. \cite[Section 2.2.3]{edelman1998geometry}. In \cite{edelman1998geometry} it is further stated that, regarding parallel transport for vectors on Riemannian homogeneous spaces, that the corresponding differential equation ``is integrable in closed form, but it is an open question whether this can be accomplished with [$O(Nn^2)$] operations.'' An interesting recent development regarding this question can be found in \cite[Proposition 4]{nguyen2024parallel}, where a relatively cheap closed form solution of parallel transport on the Stiefel manifold is derived.

For the purposes of this work we however do not use parallel transport with respect to the Levi-Civita connection, but instead use simpler expressions associated with the \textit{canonical invariant connection of second kind} \cite{nomizu1954invariant, schlarb2024covariant}:

\begin{proposition}
Given two elements $B^A_1, B^A_2\in{}T_AG$ the \textbf{parallel transport} of $B^A_2$ along the geodesic of $B^A_1$ with respect to the \textit{canonically invariant connection of second kind} is:

\begin{equation}
    \Pi_{A\to\gamma_{B^A_1}(t)}B_2^A = A\exp(t\cdot{}A^{-1}B^A_1)A^{-1}B^A_2 = A\exp(t\cdot{}B_1)B_2,
\end{equation}

where $B_i := A^{-1}B^A_i.$
\end{proposition}

For the Stiefel manifold we have \cite{schlarb2024covariant}:

\begin{proposition}
Given two elements $\Delta_1, \Delta_2\in{}T_Y\mathcal{M}$, the parallel transport of $\Delta_2$ along the geodesic of $\Delta_1$ is given by

\begin{equation}
\begin{split}
    \Pi_{Y\to\gamma_{\Delta_1}(t)}\Delta_2 = \exp(t\cdot\Omega(\Delta_1))\Delta_2 =  \lambda(Y)\exp(\mathcal{B}_1)\lambda(Y)^{-1}\Delta_2,
\end{split}
\end{equation}
where $\mathcal{B}_1 = \lambda(Y)^{-1}\Omega(\Delta_1)\lambda(Y).$
\end{proposition}

We can further modify the expression of parallel transport for the Stiefel manifold: 

\begin{equation}
\Pi_{Y\to\gamma_{\Delta_1}(t)}\Delta_2 = \lambda(Y)\exp(\mathcal{B}_1)\lambda(Y)\Omega(\Delta_2)Y = \lambda(Y)\exp(\mathcal{B}_1)\mathcal{B}_2E,
\end{equation}

where $\mathcal{B}_2 = \lambda(Y)^{-1}\Omega(\Delta_2)\lambda(Y).$ We can now define explicit updating rules for the global section $\Lambda^{(\cdot)}$ (this is the operation \texttt{update\_section} in \Cref{alg:general_man}), the element of the homogeneous space $Y^{(\cdot)}$, the tangent vector $\Delta^{(\cdot)}$ and $D^{(\cdot)} = (\Lambda^{(\cdot)})^{-1}\Omega(\Delta^{(\cdot)})\Lambda^{(\cdot)}$, its representation in $\mathfrak{g}^\mathrm{hor}$. In \Cref{alg:general_man} the geodesic direction of the $t$-th step is the final velocity $W^{(t)}\in\mathfrak{g}^{\mathrm{hor}}$, so that $\mathcal{B}^{(t)} = W^{(t)}$ in the updating rules below.

We thus have:

\begin{enumerate}
\item $\Lambda^{(t+1)} \leftarrow \Lambda^{(t)}\exp(\mathcal{B}^{(t)}) = \mathtt{update\_section}(\Lambda^{(t)}, W^{(t)}),$
\item $Y^{(t + 1)} \leftarrow \Lambda^{(t + 1)}E,$
\item $\Delta^{(t + 1)} \leftarrow  \Lambda^{(t)}\exp(\mathcal{B}^{(t-1)})^{-1}(\Lambda^{(t)})^{-1}\Delta^{(t)} = \Lambda^{(t + 1)}D^{(t)}E,$
\item $D^{(t + 1)} \leftarrow D^{(t)}.$
\end{enumerate}

So we conveniently take parallel transport of vectors into account by representing them in $\mathfrak{g}^\mathrm{hor}$: $D$ does not change (with our choice of connection), i.e. we always have $D^{(t+1)} = D^{(t)}$.

\begin{figure}
\centering
\begin{subfigure}{.475\textwidth}
\includegraphics[width = .99\textwidth]{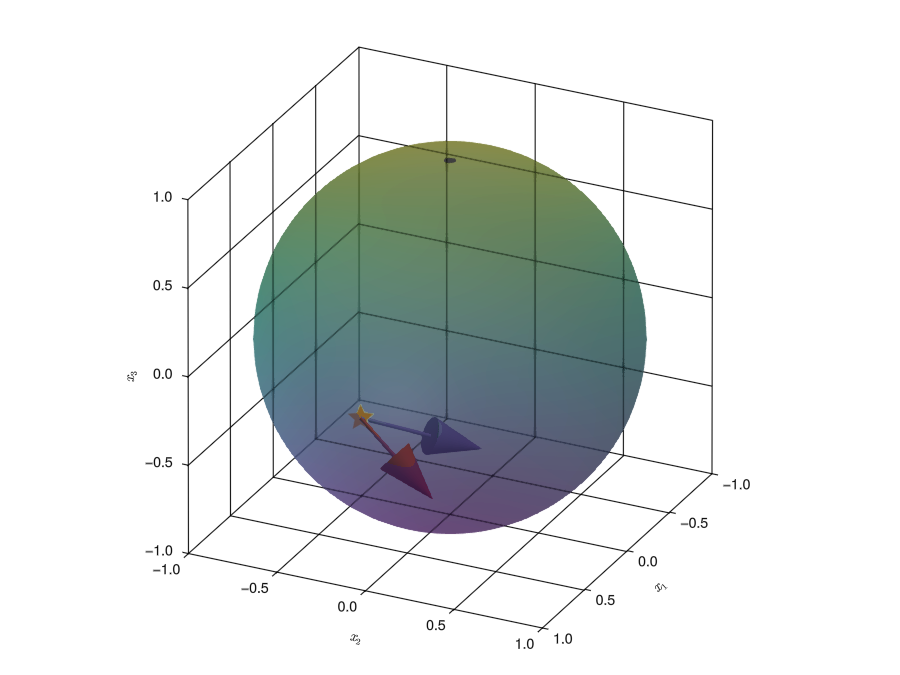}
\caption{The red vector $v_1$ represents the \textit{final velocity} and the purple vector $v_2$ represents the \textit{cache} of the optimizer.}
\label{fig:before_pt}
\end{subfigure}\hfill
\begin{subfigure}{.475\textwidth}
    \includegraphics[width = 0.99\textwidth]{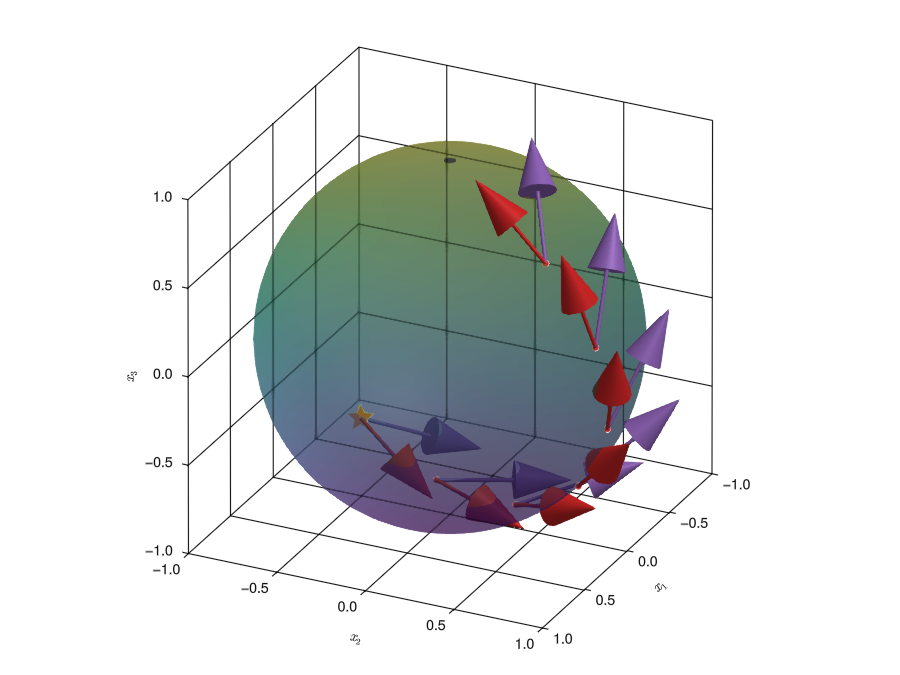}
    \caption{Here the geodesic equation has been solved for $v_1$ for different time steps $\eta$ and $v_2$ has been parallel-transported along it.}
\label{fig:after_pt}
\end{subfigure}
\caption{Visualization of parallel transport.}
\label{fig:parallel_transport}
\end{figure}

To demonstrate parallel transport we compute the geodesic along a vector and parallel transport another one along this geodesic. The result is shown in \Cref{fig:parallel_transport}. 

Here we perform parallel transport on the homogeneous space $S^2 = St(1, 3)$, i.e. the sphere. \Cref{fig:before_pt} shows the two vectors before the geodesic is solved. We then compute $\mathtt{geodesic}(\eta\cdot{}v_1)$, where $v_1$ is the red vector, for a number of steps $\eta$. The purple vector, which we call $v_2$, is parallel transported along the geodesic determined by $v_1$. When training a neural network, $v_1$ can be seen as \textit{final velocity}, i.e. the output of an optimizer method (confer \Cref{alg:general_man} and \Cref{dftn:method}), and $v_2$ can be seen as the \textit{cache} belonging to that optimizer.

We further note that parallel transport can be defined differently still, as is done for example in \cite[Section 8.1.2]{absil2008optimization}.

\section{The Softmax Activation Function and the Classification Layer}

\label{softmax}

The softmax function maps an arbitrary vector $v\in\mathbb{R}^\ell$ to a probability vector $\mathrm{softmax}(v)\in\{w\in(0,1)^\ell:\sum_{i=1}^\ell{}w_i=1\}$ via

\begin{equation}
    \mathrm{softmax}(v) = \frac{\exp(v_i)}{\sum_{i=1}^\ell\exp(v_i)}.
\end{equation}

In \Cref{transformer} we referred to a prediction of the form $(0, \ldots, 1, \ldots, 0) =: e_i$, i.e. a vector that has entry 1 at the $i$-th slot and zeros at all others, as a \textit{trivial prediction}. This is what is learned by the network in \Cref{fig:preprocessing} if we do not constrain the projection matrices to be on the Stiefel manifold. The loss of the standard Adam ablation is consistently stuck at around 1.34; this is because the output of the network is always $e_i$ and the target is a vector $t\in\{0,1\}^{10}$ with zeros everywhere except for one component. So the resulting $L_2$ error is
\begin{equation}
\sqrt{\frac{9}{10}\cdot2} \approx 1.34,
\end{equation}
which is what we observe for the standard Adam ablation in \Cref{fig:mnist_seed_band,fig:fashion_seed_band}.

\section{Computing Correlations in the Multihead-Attention Layer}
\label{mha_correlation}

In \Cref{transformer} multihead attention was described in three steps. Here we elaborate on the second one of these and argue why it makes sense to constrain the projection matrices to be part of the Stiefel manifold. 

The \textit{attention mechanism} in \Cref{eq:attention} describes a reweighting of the ``values'' $V_i$ based on correlations between the ``keys'' $K_i$ and the ``queries'' $Q_i$. First note the structure of these matrices: they are all a collection of 16 7-dimensional vectors, i.e. $V_i=[v_i^{(1)}, \ldots, v_i^{(16)}], K_i=[k_i^{(1)}, \ldots, k_i^{(16)}]$ and  $Q_i=[q_i^{(1)}, \ldots, q_i^{(16)}]$ . Those vectors have been obtained by applying the respective projection matrices onto the original image $I_i\in\mathbb{R}^{49\times16}$.

When performing the \textit{reweighting} of the columns of $V_i$ we first compute the correlations between the vectors in $K_i$ and in $Q_i$ and store the results in a \textit{correlation matrix} $C_i$: 

\begin{equation}
    [C_i]_{mn} = \left(k_i^{(m)}\right)^Tq_i^{(n)}.
    \label{eq:correlation} 
\end{equation}  

The columns of this correlation matrix are than rescaled with a softmax function, obtaining a matrix of \textit{probability vectors} $\mathcal{P}_i$:

\begin{equation}
    [\mathcal{P}_i]_{\bullet{}n} = \mathrm{softmax}([C_i]_{\bullet{}n}).
\end{equation}

Finally the matrix $\mathcal{P}_i$ is multiplied onto $V_i$ from the right, resulting in 16 convex combinations of the 16 vectors $v_i^{(m)}$ with $m=1,\ldots,16$:

\begin{equation}
    V_i\mathcal{P}_i = \left[\sum_{m=1}^{16}[\mathcal{P}_i]_{m1}v_i^{(m)}, \ldots, \sum_{m=1}^{16}[\mathcal{P}_i]_{m16}v_i^{(m)}\right].
\end{equation}

With this we can now give a better interpretation of what the projection matrices $W_i^V$, $W_i^K$ and $W_i^Q$ should do: they map the original data to lower-dimensional subspaces. We then compute correlations between the representation in the $K$ and in the $Q$ basis and use this correlation to perform a convex reweighting of the vectors in the $V$ basis. These reweighted \textit{values} are then fed into a standard feedforward neural network.

Because the main task of the $W_i^V$, $W_i^K$ and $W_i^Q$ matrices here is for them to find bases, it makes sense to constrain them onto the Stiefel manifold; they do not need to have the maximum possible generality.


\end{document}